\documentclass[lettersize,journal]{IEEEtran}
\usepackage{amsmath,amsfonts}
\usepackage{algorithmic}
\usepackage{algorithm}
\usepackage{array}
\usepackage[caption=false,font=normalsize,labelfont=sf,textfont=sf]{subfig}
\usepackage{textcomp}
\usepackage{stfloats}
\usepackage{url}
\usepackage{verbatim}
\usepackage{graphicx}
\usepackage{cite}
\usepackage[utf8]{inputenc} 
\usepackage[T1]{fontenc}    
\usepackage{hyperref}       
\usepackage{url}            
\usepackage{booktabs}       
\usepackage{amsfonts}       
\usepackage{nicefrac}       
\usepackage{microtype}      
\usepackage{xcolor}         
\usepackage{graphicx}       
\usepackage{amsmath,amsthm,amsfonts,amssymb,bm}
\usepackage{algorithm,algorithmic}
\usepackage[table,xcdraw]{xcolor}
\usepackage{setspace}
\usepackage{wrapfig}
\usepackage{graphicx}
\begin{document}

\title{DIFF-MF: A Difference-Driven Channel-Spatial State Space Model for Multi-Modal Image Fusion}

\author{Yiming~Sun, Zifan~Ye, Qinghua~Hu,~\IEEEmembership{Senior Member,~IEEE}, and Pengfei~Zhu
\thanks{Yiming~Sun, Zifan~Ye, and Pengfei~Zhu are with the School of Automation, Southeast University, Nanjing 210096, China. Pengfei~Zhu is also with the Low-Altitude Intelligence Lab, Xiong'an National Innovation Center Technology Co., Ltd. and Xiong'an Guochuang Lantian Technology Co., Ltd., Hebei 070001, China (e-mail: sunyiming@seu.edu.cn; yezifan@seu.edu.cn; zhupengfei@tju.edu.cn). \textit{(Corresponding author: Pengfei~Zhu)}
\par Qinghua~Hu are with the School of Artificial Intelligence, Tianjin University, Tianjin 300403, China (e-mail: huqinghua@tju.edu.cn).}
}

\markboth{IEEE Transactions on Neural Networks and Learning Systems}%
{Shell \MakeLowercase{\textit{et al.}}: A Sample Article Using IEEEtran.cls for IEEE Journals}

\IEEEpubid{0000--0000/00\$00.00~\copyright~2021 IEEE}

\maketitle

\begin{abstract}
Multi-modal image fusion aims to integrate complementary information from multiple source images to produce high-quality fused images with enriched content. Although existing approaches based on state space model have achieved satisfied performance with high computational efficiency, they tend to either over-prioritize infrared intensity at the cost of visible details, or conversely, preserve visible structure while diminishing thermal target salience. To overcome these challenges, we propose DIFF-MF, a novel difference-driven channel-spatial state space model for multi-modal image fusion. Our approach leverages feature discrepancy maps between modalities to guide feature extraction, followed by a fusion process across both channel and spatial dimensions. In the channel dimension, a channel-exchange module enhances channel-wise interaction through cross-attention dual state space modeling, enabling adaptive feature reweighting. In the spatial dimension, a spatial-exchange module employs cross-modal state space scanning to achieve comprehensive spatial fusion. By efficiently capturing global dependencies while maintaining linear computational complexity, DIFF-MF effectively integrates complementary multi-modal features. Experimental results on the driving scenarios and low-altitude UAV datasets demonstrate that our method outperforms existing approaches in both visual quality and quantitative evaluation. Our code will be available at \href{https://github.com/ZifanYe-SEU/DIFF_MF}{https://github.com/ZifanYe-SEU/DIFF\_MF}.
\end{abstract}

\begin{IEEEkeywords}
Infrared-visible image fusion, difference-driven, channel-exchange, spatial-exchange, cross-modal state space scanning.
\end{IEEEkeywords}

\begin{figure*}[t]
	\centering
	\includegraphics[width=\linewidth]{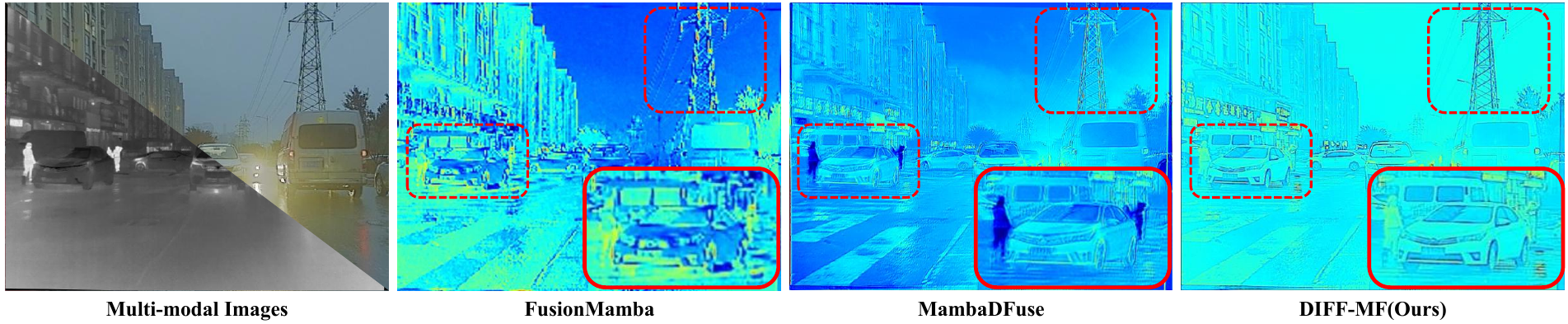}
	\caption{The feature visualization of different Mamba-based image fusion models. FusionMamba's over-fused features are dominated by infrared intensity and MambaDFuse's lack of interaction fails to highlight the thermal signatures of pedestrians. Our DIFF-MF effectively preserves thermal signatures from infrared modality and texture details from visible modality while maintaining optimal edge textures in background regions.}
	\label{fig1}
\end{figure*}

\section{Introduction}
Multimodal image fusion aims to generate high-quality fused images that comprehensively represent scene contents by learning complementary information from two modalities, thereby enhancing the performance of downstream tasks like object detection and semantic segmentation~\cite{10177917,10278227,9697423}. Taking infrared-visible image fusion as an example, infrared images provide illumination-independent thermal imaging capabilities that effectively compensate for visible images' nighttime perception limitations, yet they typically lack texture information for intuitive identification of thermal targets. Visible images exhibit superior imaging quality with rich texture and color details under adequate illumination. This fusion technique has found widespread applications in scenarios like UAV urban inspections and disaster rescue operations.

Recent years have witnessed growing attention on effectively integrating complementary features from different modalities. Various methods have been proposed, including pre-trained CNN-based approaches~\cite{li2018densefuse,zhang2020ifcnn,li2020nestfuse,zhao2023interactive}, autoencoder architectures, GAN frameworks~\cite{liu2022target}, diffusion models~\cite{zhao2023ddfm,yi2024diff,yang2025dsfuse}, and Transformer~\cite{zhao2021dndt,ma2022swinfusion,qu2022transmef,vs2022image,fu2021ppt} networks. However, these methods face inherent limitations: CNN-based approaches suffer from restricted receptive fields hindering global context capture; autoencoders inherit CNN's defects while struggling with multimodal complementarity modeling; GANs encounter training instability leading to adversarial learning collapse; diffusion models suffer from inefficient inference due to prolonged training and sampling requirements; Transformer architectures, despite superior global perception, demand excessive computational resources due to quadratic complexity of attention mechanisms. Notably, the emerging Mamba architecture enables global information modeling with linear computational complexity.
\IEEEpubidadjcol

While recent Mamba-based fusion methods like FusionMamba~\cite{xie2024fusionmamba} and MambaDFuse~\cite{li2024mambadfuse} have emerged to address computational efficiency, they seem to suffer from a cross-modal information imbalance, typically exhibiting a bias towards a single dominant modality. Specifically, FusionMamba~\cite{xie2024fusionmamba} fuses multimodal features before SSM learning, potentially diluting modality-specific information. In contrast, MambaDFuse~\cite{li2024mambadfuse} processes each modality independently via SSMs and then fused the features with a static mean-map strategy. This may restricts effective token-level interaction, risking the loss of contextual information. As illustrated in Fig.~\ref{fig1}, FusionMamba's over-fused features are dominated by infrared intensity, weakening the visible texture details of vehicles; meanwhile, MambaDFuse's lack of interaction fails to highlight the thermal signatures of pedestrians. Unlike existing fusion methods, our method leverages feature discrepancy maps as a feature extraction guidance. This structure forces the network to focus on the unique variations between modalities. Moreover, the channel-wise interaction and spatial token-mixing scanning mechanisms will adequately fuse the cross-modal features, effectively preserving thermal signatures from infrared modality and texture details from visible modality while maintaining optimal edge textures in background regions.

This paper proposes a difference-driven channel-spatial state space model for multimodal image fusion, termed DIFF-MF. The framework is built upon a multimodal feature extraction backbone, enhanced by three dedicated modules: a difference guidance module; a channel-exchange module; and a spatial-exchange module. Specifically, following feature extraction by the backbone, the framework employs a difference guidance module to perform feature weighting, while a channel-exchange module conducts cross-modal, channel-reweighted state-space exchange for calibration. For spatial processing, we propose three scanning mechanisms operating at different feature scales to capture multiscale spatial information and global cross-modal interactions while maintaining linear complexity. This design enables DIFF-MF to effectively model both channel-wise and spatial global contexts, preserving both modality-specific and cross-modal common features. Our main contributions are summarized as follows: 
\begin{itemize}
\item We propose a novel differential-driven channel-spatial SSM, which uniquely models both modality-specific differential features and cross-modal collaborative global contexts, leading to superior fusion performance.
\item We propose a mechanism for differential feature recalibration and cross-modal channel-exchange reweighting, which balances modality dominance and feature importance through channel-level calibration, thereby achieving equitable representation of multimodal features.
\item We design multi-scale cross-modal spatial scanning mechanisms, enabling comprehensive global interaction across feature scales, with extensive experiments validating our method's state-of-the-art performance in fusion and downstream tasks.
\end{itemize}

The remainder of the paper is organized as follows:
In Section~\ref{sec:2}, we present related work. 
In Section~\ref{sec:3}, we present our DIFF-MF in detail. 
We describe extensive experiments in Section~\ref{sec:4} and conclude the paper in Section~\ref{sec:5}.

\section{Related Works}
\label{sec:2}
\subsection{Multi-Modal Image Fusion}
Multi-modal image fusion aims to integrate complementary information from different imaging sensors into a single composite image that is more informative and suitable for human perception or subsequent computer vision tasks~\cite{10190200,10026659,10713288}. With the advancement of deep learning, a variety of data-driven fusion strategies have been proposed, which can be broadly categorized into the following types:

\textbf{CNN-based methods} constitute one of the earliest and most widely adopted deep learning approaches for image fusion. These methods typically employ convolutional encoders to extract multi-scale features from source images, followed by fusion rules applied in the feature domain. Representative works include DenseFuse~\cite{li2018densefuse}, which uses dense connections to enhance feature reuse, and IFCNN~\cite{zhang2020ifcnn}, a general fusion framework that leverages adaptive fusion strategies. U2Fusion~\cite{Xu2022U2FusionAU} introduced an unsupervised framework that optimizes the similarity between the fused image and source images without requiring ground-truth labels. DIDFuse~\cite{zhao2020didfuse} combined CNN with image decomposition to separate base and detail layers for more robust fusion. Despite their efficiency and ease of training, CNN-based methods are inherently limited by local receptive fields, which restrict their ability to capture long-range dependencies and global contextual information, often leading to suboptimal fusion in complex scenes.

\textbf{Autoencoder-based frameworks} often leverage encoder-decoder architectures to learn compact representations of source images. DeepFuse~\cite{ram2017deepfuse} was among the first to adopt such a structure for image fusion. NestFuse~\cite{li2020nestfuse} incorporated nested connections and attention mechanisms to enhance feature reuse and highlight salient regions. RFN-Nest~\cite{li2021rfn} further improved feature extraction through residual fusion networks. While autoencoders can effectively compress and reconstruct images, they often inherit the locality constraints of CNNs and may struggle to model complex cross-modal interactions, especially when modalities exhibit significant disparities.

\textbf{Generative Adversarial Networks (GANs)} have been explored to generate fused images with enhanced realism and perceptual quality. FusionGAN~\cite{ma2019fusiongan} use GANs in infrared-visible fusion by employing an adversarial loss to encourage realistic outputs. Subsequent variants like MEF-GAN~\cite{xu2020mef} and MFF-GAN~\cite{zhang2021mff} extended the framework to multi-exposure and multi-focus fusion tasks. TarDAL~\cite{liu2022target} introduced a target-aware dual adversarial learning framework that aligns fusion with downstream detection tasks. However, GAN-based methods are prone to training instability and mode collapse, which can lead to artifacts or loss of detail in the fused output.

\textbf{Transformer-based methods} address the limitations of local modeling by leveraging self-attention mechanisms to capture global dependencies. SwinFusion~\cite{ma2022swinfusion} integrated Swin Transformer blocks into a hierarchical fusion network, enabling cross-scale long-range interaction. CDDFuse~\cite{zhao2023cddfuse} combined Transformer with correlation-driven feature decomposition to better separate common and unique features across modalities. EMMA~\cite{zhao2024equivariant} adopted a U-shaped architecture built on Restormer~\cite{zamir2022restormer} blocks for effective multi-modal fusion. Despite their superior ability to model global contexts, Transformers suffer from quadratic computational complexity, making them less efficient for high-resolution images, which limits their practicality in real-time applications.

\textbf{Diffusion models} have recently emerged as a powerful generative approach for image fusion. DDFM~\cite{zhao2023ddfm} employed a denoising diffusion process to iteratively refine the fused image, achieving high-quality results. Diff-IF~\cite{yi2024diff} incorporated a fusion knowledge prior to address the lack of ground truth in diffusion training. While diffusion models can achieve high fusion performance, their inference is computationally intensive due to multi-step sampling, hindering deployment in time-sensitive scenarios.

Moreover, there are some other models for image fusion tasks. Huang~\textit{et al.}~\cite{huang2025t} proposed a target-aware image fusion network leveraging Taylor expansion approximation to decompose the images. Li~\textit{et al.}~\cite{li2025graph} adopted graph convolution networks(GCNs)~\cite{micheli2009neural} to integrate the graph representation into the infrared and visible fusion. While existing methods have made significant progress, they often trade off between global modeling capability and computational efficiency. Our proposed DIFF-MF seeks to achieve a balance by incorporating a difference-driven dual-branch architecture with channel-spatial state space modules, enabling efficient and effective fusion of multi-modal images.

\begin{figure*}[!t]
	\centering
	\includegraphics[width=1.0\linewidth]{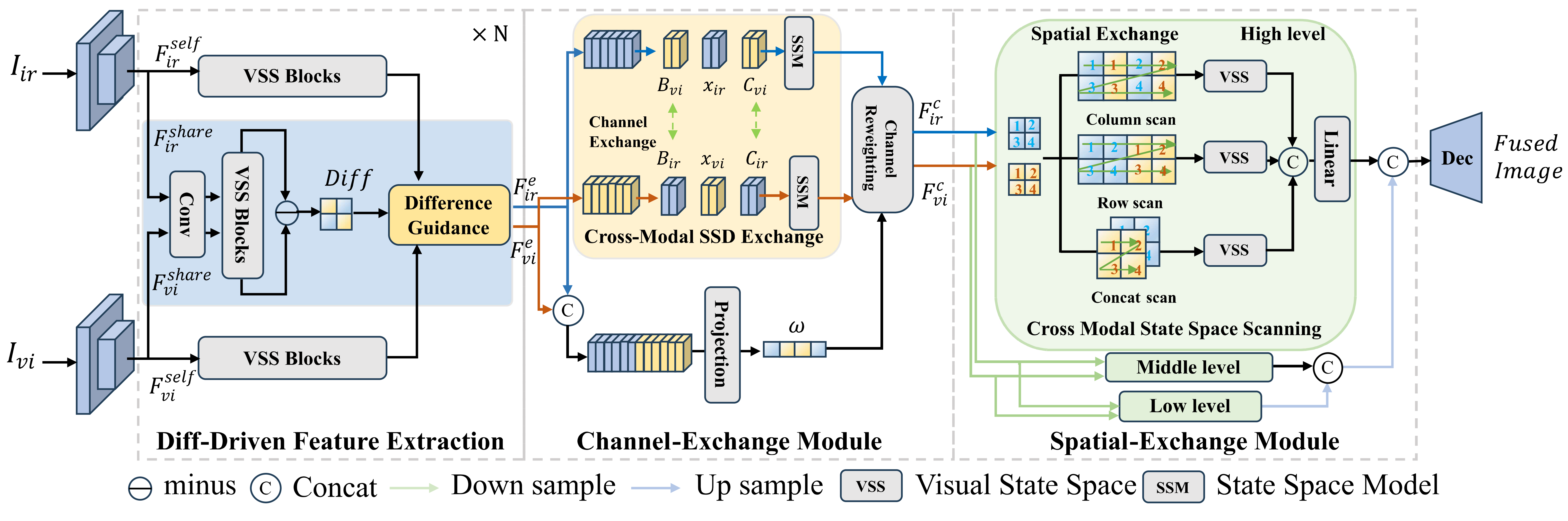} 
	\caption{The architecture of DIFF-MF. DIFF-MF consists of a difference driven feature extraction , a channel-exchange module, and a spatial-exchange module.}
	\label{fig:Framework}
\end{figure*}

\subsection{State Space Model}
State Space Models(SSMs)\cite{gu2021combining}, developed from modern control theory, can be considered as linear time-invariant(LTI) systems. SSMs are able to capture long-range information by mapping the 1D input \(x(t) \in \mathbb{R}^N\) to the output \(y(t) \in \mathbb{R}^N\) through a hidden state \(h(t) \in \mathbb{R}^N\). This process can be formulated as the following linear ordinary differential equations(ODEs):
\begin{equation}
\begin{gathered}
h^{\prime}(t) = \bm{A}h(t) + \bm{B}x(t),\\
y(t)  = \bm{C}h(t) + \bm{D}x(t),
\end{gathered}
\end{equation}
where \(\bm{A} \in \mathbb{R}^{N \times N}\), \(\bm{B} \in \mathbb{R}^{N \times 1}\), \(\bm{C} \in \mathbb{R}^{N \times 1}\), and \(\bm{D} \in \mathbb{R}^{N}\) can be regarded as projection matrices.
To combined the SSMs with deep learning, the discretization process is applied to Eq. (1) through a timescale parameter \(\Delta\) which can discretize the projection matrices\(\bm{A} \in \mathbb{R}^{N \times N}\), \(\bm{B} \in \mathbb{R}^{N \times 1}\) to discrete forms \(\overline{\bm{A}}\), \(\overline{\bm{B}}\). Commonly, the zero-order hold rule(ZOH) is used to complete the discretization process. It can be formulated as:
\begin{equation}
\begin{gathered}
\overline{\bm{A}} = \exp\left( \Delta\bm{A} \right),\\
\overline{\bm{B}} = \left( \Delta \bm{A} \right)^{-1} \left( \exp\left( \Delta \bm{A} \right) - \bm{I} \right) \cdot \Delta \bm{B}.
\end{gathered}
\end{equation}
Then Eq. (1) can be written in discrete form as follows:
\begin{equation}
\begin{gathered}
h_{k}=\overline{\bm{A}}h_{k-1}+\overline{\bm{B}}x_{k},\\
y_{k}=\bm{C}h_{k}+\bm{D}x_{k}.
\end{gathered}
\end{equation}
Based on Eq. (3), Mamba~\cite{gu2023mamba} introduces a selective scan mechanism that turns \(\Delta\), \(\bm{B}\), \(\bm{C}\) into input-dependent parameters.

The state space model can be interpreted as a dual form of the self-attention mechanism in Transformer blocks, constituting a special case of generalized attention~\cite{dao2024transformers}.
Models defined through a linear-time recurrence usually can derive a quadratic form by expanding the matrix formulation characterizing its linear sequence-to-sequence transformation, while models defined through quadratic-time pairwise interactions can derive a linear form by viewing it as a four-way tensor contraction and reducing in a different order.
This is called Structured State Space Duality (SSD)~\cite{dao2024transformers} where the \(\bm{C}\), \(\bm{B}\), \(x\) in the SSM are corresponding to the \(\bm{Q}\), \(\bm{K}\), \(\bm{V}\) in the self-attention mechanism. Then, a dual form of cross-attention mechanism can be developed according to the SSD theory.

\subsection{Mamba-based Image Fusion}
Due to its ability to efficiently model long-range dependencies, Mamba~\cite{gu2023mamba} has been widely adopted and has opened new research avenues in computer vision. Unlike Transformers, which rely on self-attention mechanisms with quadratic complexity, Mamba leverages state space models (SSMs) to capture global context with linear computational complexity. This makes it particularly suitable for high-resolution image processing tasks such as multi-modal image fusion, where both efficiency and global receptive fields are critical.

Several pioneering works have adapted Mamba for vision tasks. Vim~\cite{zhu2024vision} applied mamba to the Vision Transformer(ViT)~\cite{han2022survey}, developing a visual backbone based on bidirectional Mamba
blocks. VMamba~\cite{liu2024vmamba} introduced a cross-scan mechanism within its SS2D module to adapt SSMs to 2D image data, allowing effective spatial context aggregation without the computational burden of self-attention. Subsequent studies have further extended Mamba’s application to various vision domains: Wang~\textit{et al.}~\cite{wang2024mamba} integrated SSMs into object detection frameworks, while Xing~\textit{et al.}~\cite{xing2024segmamba} applied Mamba to medical image segmentation, demonstrating its capability in capturing long-range spatial dependencies. Lin~\textit{et al.}~\cite{lin2024mtmamba} proposed MTMamba for dense scene understanding, highlighting Mamba’s versatility in multi-task learning scenarios.

In the specific domain of multi-modal image fusion, two recent Mamba-based approaches have emerged: FusionMamba~\cite{xie2024fusionmamba} and MambaDFuse~\cite{li2024mambadfuse}. FusionMamba employs an efficient SS2D module combined with a cross-modality fusion block to integrate features from different modalities. However, it performs feature fusion before SSM-based processing, which may lead to the dilution of modality-specific characteristics early in the pipeline. On the other hand, MambaDFuse introduces an M3 block to process each modality separately through SSMs, followed by a mean-map-weighted fusion strategy. Although this preserves modality-specific information to some extent, it lacks explicit cross-modal interaction during token-level processing, potentially limiting the model’s ability to capture synergistic features.

Despite these advances, existing Mamba-based fusion methods have not fully exploited the potential of SSMs in modeling differential features and facilitating cross-modal token mixing. Effective multi-modal fusion requires not only global context aggregation but also careful handling of modality-specific discrepancies and complementary information. These limitations underscore the need for an approach that explicitly leverages differential guidance and cross-modal state-space interactions.

Our proposed DIFF-MF addresses these gaps by introducing a difference-driven channel-spatial state space model. Unlike prior works, DIFF-MF uses feature discrepancy maps to guide both channel-wise and spatial-wise fusion processes. The channel-exchange module enables dynamic cross-modal reweighting via a dual SSM formulation, while the spatial-exchange module employs multi-scale scanning mechanisms to ensure thorough global integration. By maintaining linear complexity and enabling explicit differential feature modeling, DIFF-MF achieves a better balance between efficiency and fusion quality, as validated by extensive experiments in Section~\ref{sec:4}.

\section{Methods}
\label{sec:3}
\subsection{Overall Architecture}
In this paper, we propose an image fusion framework termed DIFF-MF. In Fig.~\ref{fig:Framework}, DIFF-MF mainly contains three modules: a difference-driven feature extraction, a channel-exchange module, and a spatial-exchange module. Firstly, we send a pair of infrared image \(I_{ir}\in \mathbb{R}^{\mathrm H\times \mathrm W\times1}\) and visible image \(I_{vi}\in \mathbb{R}^{\mathrm H\times \mathrm W\times3}\) into the difference-driven feature extraction module to extract features respectively. Then the extracted features, denoted as \({F}_{vi}^{e}\) and \({F}_{ir}^{e}\), are sent into the channel-exchange module to fuse the features in the channel dimension. After that, the fused features, denoted as  \({F}_{vi}^{c}\) and \({F}_{ir}^{c}\), will enter the spatial-exchange module, which will fuse the features in the spatial dimension. At last, the feature maps fused by the spatial-exchange module can be decoded into the final fused image. 

\begin{figure}[h]
	\centering
	\includegraphics[width=\linewidth]{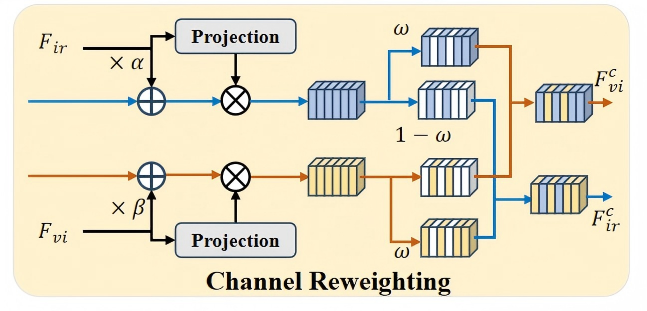} 
	\caption{The architecture of channel reweighting. Parameters \(\alpha,\beta\) are learned from the \({F}_{vi}^{e}\) and \({F}_{ir}^{e}\). \(\omega\) is generated by the Gate Generator.}
	\label{fig:reweight}
\end{figure}

\subsection{Diff-Driven Feature Extraction}
The input features \(I_{ir}\) and \(I_{vi}\) are processed by two distinct types of dual branches: weight-dependent branches and weight-shared branches. As shown in Fig.~\ref{fig:Framework}, the weight-dependent dual branches consist of Visual State Space (VSS) blocks~\cite{liu2024vmamba} while the weight-shared branches contain CNN blocks and VSS blocks, both of them are weight-shared. Then the output feature maps of weight-dependent dual branches are sent into the difference guidance module to re-weight the feature maps through a difference mask generated by the output of weight-shared dual branches. Let \({F}_{vi}^{self}\), \({F}_{ir}^{self}\) represent the features of the weight-dependent dual branches and \({F}_{vi}^{share}\), \({F}_{ir}^{share}\) represent the features of the weight-shared branches, the entire feature extraction process can be described in Algorithm~\ref{alg:Algorithm1}.

\begin{algorithm}
\scriptsize
\setstretch{1.5}
    \caption{\bf Diff Driven Feature Extraction}
    \label{alg:Algorithm1}
    \renewcommand{\algorithmicrequire}{\textbf{Input:}}
    \renewcommand{\algorithmicensure}{\textbf{Output:}}
    \begin{algorithmic}[1]
        \REQUIRE ${{F}_{vi}^{self^n}, {F}_{ir}^{self^n},{F}_{vi}^{share^n},{F}_{ir}^{share^n}}\mathrm{\textcolor[RGB]{0, 100, 0}{(B,H,W,C)}}$
        \ENSURE ${{F}_{vi}^{self^{n+1}}, {F}_{ir}^{self^{n+1}},{F}_{vi}^{share^{n+1}},{F}_{ir}^{share^{n+1}}}\mathrm{\textcolor[RGB]{0, 100, 0}{(B,H,W,C)}}$
        \STATE ${{F}_{vi}^{self^n}} \gets {VSS({F}_{vi}^{self^n})}\mathrm{\textcolor[RGB]{0, 100, 0}{(B,H,W,C)}}$
        \STATE ${{F}_{ir}^{self^n}} \gets {VSS({F}_{ir}^{self^n})}\mathrm{\textcolor[RGB]{0, 100, 0}{(B,H,W,C)}}$
        \STATE ${{F}_{vi}^{share^{n+1}},{F}_{ir}^{share^{n+1}}} \gets {VSS(Conv({F}_{vi}^{share^n},{F}_{ir}^{share^n}))}\mathrm{\textcolor[RGB]{0, 100, 0}{(B,H,W,C)}}$
        \STATE ${Diff} \gets {Tanh|{F}_{vi}^{share^{n+1}}-{F}_{ir}^{share^{n+1}}|}\mathrm{\textcolor[RGB]{0, 100, 0}{(B,H,W,C)}}$
        \STATE ${{F}_{ir}^{self^{n+1}}} \gets {(1-Diff)\odot{F}_{ir}^{self^n}+Diff\odot{{F}_{vi}^{self^n}}}\mathrm{\textcolor[RGB]{0, 100, 0}{(B,H,W,C)}}$
        \STATE ${{F}_{vi}^{self^{n+1}}} \gets {(1-Diff)\odot{F}_{vi}^{self^n}+Diff\odot{{F}_{ir}^{self^{n+1}}}}\mathrm{\textcolor[RGB]{0, 100, 0}{(B,H,W,C)}}$
        \RETURN ${{F}_{vi}^{self^{n+1}}, {F}_{ir}^{self^{n+1}},{F}_{vi}^{share^{n+1}},{F}_{ir}^{share^{n+1}}}\mathrm{\textcolor[RGB]{0, 100, 0}{(B,H,W,C)}}$        
    \end{algorithmic}
\end{algorithm}

\subsection{Channel-Exchange Module}
Channel-exchange module consists of two parts: the cross-modal State Space Duality (SSD) exchange module and the gated-control channel reweighting module. Inspired by~\cite{dao2024transformers}, the cross-modal SSD exchange module, as shown in Fig~\ref{fig:Framework}, is designed to exchange the channels between the infrared modality and visible modality based on the theories of Structured State Space Duality and SSMs. This exchange method is integrated into the cross-scan mechanism of the VSS blocks. Let \({F}_{vi}^{e}\) and \({F}_{ir}^{e}\) represent the input feature maps. Consider one route scan as an example. Its first step is formulated as follows:
\begin{equation}
\begin{gathered}
{{T}_{ir}} = {CrossScan}({{F}_{ir}^{e}}),\\
{{T}_{vi}} = {CrossScan}({{F}_{vi}^{e}}),
\end{gathered}
\end{equation}
where \({T}_{ir}\in \mathbb{R}^{\mathrm B \times \mathrm L\times \mathrm C}\) and \({T}_{vi}\in \mathbb{R}^{\mathrm B \times \mathrm L\times \mathrm D}\) denote the tokens generated from the cross scan. Here \(\mathrm{B}\) means the batch size, \(\mathrm{L}\) means the sequence length and \(\mathrm{C}\) means the channels.
Then the tokens are projected into a new channel size for channel exchange and split into \(x\), \({B}\), \({C}\) which are the dual form of \(\mathbf{Q}\), \(\mathbf{K}\), \(\mathbf{V}\) in the structured mask attention\cite{dao2024transformers}. It can be represent by the following equation:
\begin{equation}
\begin{gathered}
x_{{ir}},{{B_{ir}}},{{C_{ir}}}={{Split}}({{Projection}}({{T}_{ir}})),\\
x_{{vi}},{{B_{vi}}},{{C_{vi}}}={{Split}}({{Projection}}({{T}_{vi}})).
\end{gathered}
\end{equation}

After that, these tokens are exchanged and reprojected into the original channel size and then sent into SSMs to learn the fused features. Finally, the output tokens are merged to the original shape:
\begin{equation}
\begin{gathered}
{{F}_{vi}^{e'}}={{CrossMerge}}({ SSM}({ Projection}[x_{{vi}},{{B_{ir}}},{{C_{ir}}}]),\\
{{F}_{ir}^{e'}}={{CrossMerge}}({ SSM}({ Projection}[x_{{ir}},{{B_{vi}}},{{C_{vi}}}]),
\end{gathered}
\end{equation}
where \({F}_{vi}^{e'}\) and \({F}_{ir}^{e'}\) denote the outputs of the cross-modal SSD exchange module. Then the outputs will be reweighted by the weight parameter \(\omega \in \mathbb{R}^{\mathrm B\times \mathrm C \times 1}\), which is generated by the gate generator, as shown in Fig.~\ref{fig:reweight}. This process can be separated into two parts, the first step can be described as follows:
\begin{equation}
\label{eq:res}
\begin{gathered}
{{F}_{ir}^{e''}}=({{{F}_{ir}^{e’}}}+{{{F}_{ir}^{e}}}\times \alpha)\odot{{{Projection}}({{{F}_{ir}^{e’}}})},\\
{{F}_{vi}^{e''}}=({{{F}_{vi}^{e’}}}+{{{F}_{vi}^{e}}}\times \beta)\odot{{{Projection}}({{{F}_{vi}^{e’}}})},\\
\end{gathered}
\end{equation}
where \(\alpha,\beta\) are learnable parameters learned from \({F}_{vi}^{e}\) and \({F}_{ir}^{e}\), then the temporary features will enter the next reweighting process formulated as follows:
\begin{equation}
\begin{gathered}
{\omega}={{Projection}}({ Concat}({{F}_{ir}^{e}},{F}_{ir}^{e})),\\
{{{F}_{ir}^{c}}}=(1-\omega)\odot{{F}_{ir}^{e''}}+{\omega }\odot{{F}_{vi}^{e''}},\\
{{{F}_{vi}^{c}}}=(1-\omega)\odot{{F}_{vi}^{e''}}+{\omega} \odot{{F}_{ir}^{e''}}.
\end{gathered}
\end{equation}

\subsection{Spatial-Exchange Module}
We employs multi-scale scanning mechanisms in the spatial-exchange module to ensure thorough global integration. Spatial-exchange module consists of three cross-modal State Space Scanning blocks from high-scale level to low-scale level. Inspired by~\cite{chen2024changemamba}, we realign the feature maps into fused feature maps and then send them to the VSS blocks. Consider the feature maps \({ M_{ir}} \in \mathbb{R}^{B\times C\times H\times W}\) and \({ M_{vi}} \in \mathbb{R}^{\mathrm B\times \mathrm C\times \mathrm H\times \mathrm W}\):
\begin{equation}
\begin{gathered}
{ M_{ir}}=[{ M_{ir}^{1}},{ M_{ir}^{2}},...,{ M_{ir}^{C}}],\\
{ M_{vi}}=[{ M_{vi}^{1}},{ M_{vi}^{2}},...,{ M_{vi}^{C}}].
\end{gathered}
\end{equation}
The realignment process varies in three different types as follows:
\begin{equation}
\begin{gathered}
{ M_{f}^{column}}=[{ M_{ir}^{1}},{ M_{vi}^{1}},{ M_{ir}^{2}},{ M_{vi}^{2}},...,{ M_{ir}^{C}},{ M_{vi}^{C}}],\\
{ M_{f}^{row}}=[{ M_{ir}^{1}},{ M_{ir}^{2}},...,{ M_{ir}^{C}},{ M_{vi}^{1}},{ M_{vi}^{2}},...,{ M_{vi}^{C}}],\\
{ M_{f}^{concat}}={{Concat}}({ M_{ir}},{ M_{vi}}).
\end{gathered}
\end{equation}

Then the three fused feature maps will enter the VSS blocks and concatenate into one feature map \({ M_{f}} \in \mathbb{R}^{\mathrm B\times \mathrm 3C\times \mathrm H\times \mathrm W}\). After that, it will be projected into the original channel size of \(\bf C\) and fused with other feature maps from another two scale levels. 

\subsection{Loss Function}
The loss function follows~\cite{ma2022swinfusion}, using SSIM loss $\mathcal{L}_{ssim}$, texture loss $\mathcal{L}_{text}$, and intensity loss $\mathcal{L}_{int}$ to constrain the fusion network.
The SSIM loss~\cite{wang2004image} is defined as:
\begin{equation}
\mathcal{L}_{ssim}=w_{1}\cdot(1-ssim(I_{f},I_{ir}))+w_{2}\cdot(1-ssim(I_{f},I_{vi})),
\end{equation}
here $ssim(\cdot)$ is the  structural similarity operation, and $w_{1}, w_{2}$ are both set as $0.5$. SSIM loss is employed to constrain the structural similarity between fused image and source images.

The texture loss can guide the fusion network to preserve as many texture details as possible and is formulated as follows:
\begin{equation}
\mathcal{L}_{text}=\frac{1}{HW}\left\|\left|\nabla I_f\right|-\max(\left|\nabla I_{ir}\right|,\left|\nabla I_{vi}\right|)\right\|_1,
\end{equation}
where $\nabla$ indicates the Sobel gradient operator, which could measure texture information. $|\cdot|$ represents absolute operation, $||\cdot||_1$ the $l_1$-norm, and max($\cdot$) refers to the element-wise maximum selection.

The intensity loss can guide the fusion network to capture proper intensity information, it can be defined as follows:
\begin{equation}
\mathcal{L}_{int}=\frac{1}{HW}\left\| I_f-M( I_{ir}, I_{vi})\right\|_1,
\end{equation}
where $M(\cdot)$ is an element-wise aggregation operation, which is associated with the specific fusion scenario.

The whole loss function of the network is a weighted sum of all sub-loss function:
\begin{equation}
\mathcal{L}_{total}=\lambda_1\mathcal{L}_{ssim}+\lambda_2\mathcal{L}_{text}+\lambda_3\mathcal{L}_{int},
\end{equation}
where $\lambda_1, \lambda_2$ and $\lambda_3$ are the hyper-parameters that control the trade-off of each sub-loss function.

\section{Experiments}
\label{sec:4}
\subsection{Experimental Setting}
\noindent \textbf{Implementation Details.}
We performed experiments on a computing platform with two NVIDIA A40 GPUs. We used Adam Optimization to update the overall network parameters with the learning rate of $2.0\times 10^{-5}$. The training epoch is set to $46$ and the batch size is $4$. 

\noindent \textbf{Datasets and Partition Protocol.} We conducted experiments on three publicly available datasets: (M$^{3}$FD~\cite{liu2022target}, TNO~\cite{toet2012progress} and DroneVehicle~\cite{sun2020drone}).

{\bf M$^{3}$FD:} It contains $4,200$ infrared-visible image pairs captured by on-board cameras. We used $3,900$ pairs of images for training and the remaining $300$ pairs for evaluation.

{\bf DroneVehicle:} It contains $28,439$ infrared-visible image pairs captured by UAV cameras. We used $300$ image pairs for evaluation. 

{\bf TNO:} We performed evaluation on $55$ image pairs. 

Note that we test our models on the DroneVehicle and TNO Datasets without retraining.

\noindent \textbf{Evaluation Metrics.}
We evaluated the performance of the proposed method based on qualitative and quantitative results. The qualitative evaluation is mainly based on the visual effect of the fused image. A good fused image needs to have complementary information from multi-modal images.
The quantitative evaluation mainly uses quality evaluation metrics to measure the performance of image fusion. We selected $6$ popular metrics, including the entropy (EN)~\cite{Roberts2008AssessmentOI}, spatial frequency (SF)~\cite{Eskicioglu1995ImageQM}, standard deviation (SD), mutual information (MI)~\cite{Qu2002InformationMF}, visual information fidelity (VIF)~\cite{Han2013ANI}, and average gradient (AG)~\cite{Cui2015DetailPF}. We also evaluate the performance of the different methods on the typical downstream task, infrared-visible object detection and semantic segmentation.

\begin{table}[!t]
    \centering
    \caption{Quantitative comparison of our DIFF-MF with 7 state
of-the-art methods. \textcolor{red}{Red} indicates the best, \textcolor{blue}{blue} indicates
 the second best, and \textcolor{cyan}{cyan} indicates the third best.}
    \label{tab:compare}
    \resizebox{1\linewidth}{!}{
    \begin{tabular}{lcccccc}
        \toprule
        \multicolumn{7}{c}{Datasets: M$^{3}$FD  Dataset}\\
        Methods & EN↑ & SD↑ & SF↑ & MI↑ & VIF↑ &  AG↑  \\
        \midrule
        SwinFusion & 6.79	&	35.84	&	13.685	&	2.882		&	\textcolor{cyan}{0.774}		&	4.603	\\
        TarDAL & \textcolor{blue}{6.99}  &  \textcolor{blue}{39.1}  &  12.649 & 2.171  &  0.598 &    4.256  \\
        DIDFuse & 6.77	&	34.23	&	12.937	&	\textcolor{cyan}{2.889}		&	0.737	&		4.435	\\
        CDDFuse & 6.9	&	37.24	&	\textcolor{cyan}{14.776}	&	2.709		&	\textcolor{blue}{0.793}		&	\textcolor{cyan}{4.862}	\\
        MambaDFuse & 6.77	&	34.23	&	12.937	&	\textcolor{cyan}{2.889}	&		0.737	&	4.435	\\
        FusionMamba & 6.85	&	37.31	&	9.278	&	\textcolor{red}{3.295}	&		0.562	&	3.393	\\
        EMMA & \textcolor{cyan}{6.92}	&	\textcolor{cyan}{38.25}	&	\textcolor{blue}{15.227}	&	2.644	&	0.769 & \textcolor{blue}{5.328}	\\
        
        \bf{DIFF-MF} & \textcolor{red}{7.19} & \textcolor{red}{55.09} & \textcolor{red}{18.787} & \textcolor{blue}{3.04} & \textcolor{red}{0.825} & \textcolor{red}{6.33}\\
        
        \midrule
         \multicolumn{7}{c}{Datasets: TNO  Dataset}\\
         Methods & EN↑ & SD↑ & SF↑ & MI↑ & VIF↑ &  AG↑  \\
         \midrule
         SwinFusion & 6.97	&	41.34	&	11.485	&	2.318	&		\textcolor{cyan}{0.74}	&	4.424	\\
         TarDAL & 7.01	&	43.52	&	11.373	&	2.05	&		0.612	& 4.016\\
         DIDFuse & \textcolor{blue}{7.18}	&	\textcolor{blue}{50.42}	&	\textcolor{cyan}{12.327}	&	1.761	&	0.633	&	4.594	\\
         CDDFuse & \textcolor{cyan}{7.15}	&	46.76	&	\textcolor{blue}{13.229}	&	2.222	&		\textcolor{blue}{0.781}	&	\textcolor{cyan}{4.775}	\\
         MambaDFuse & 6.98	&	40.49	&	10.836	&	\textcolor{cyan}{2.417}	&		0.739	&	4.197	\\
         FusionMamba & 7.04	&	47.09	&	9.419	&	\textcolor{red}{7.656}	&		0.658 & 3.707	 \\
         EMMA & \textcolor{red}{7.25} &	\textcolor{cyan}{48.23}	&	12.126	&	2.062	&		0.695	&	\textcolor{blue}{4.928}	\\
        
        \bf{DIFF-MF} & 7.1	&	\textcolor{red}{63.03}	&	\textcolor{red}{14.672}	&	\textcolor{blue}{2.521}	&		\textcolor{red}{0.836}	&	\textcolor{red}{5.393}\\
        \midrule
         \multicolumn{7}{c}{Datasets: DroneVehicle  Dataset}\\
         Methods & EN↑ & SD↑ & SF↑ & MI↑ & VIF↑ &  AG↑  \\
         \midrule
         SwinFusion & 7.4	&	49.59	&	19.226	&	2.072	&		0.623	&	\textcolor{blue}{7.286}	\\
         TarDAL & 7.27	&	46.14	&	15.016	&	1.926	&	0.485	&	5.323	\\
         DIDFuse & 7.19	&	\textcolor{blue}{57.39}	&	\textcolor{blue}{19.533}	&	1.852	&	0.505	&	6.776\\
         CDDFuse & 7.36	&	46.73	&	\textcolor{cyan}{19.523}	&	2.172	&	\textcolor{blue}{0.642}	&	6.857	\\
         MambaDFuse & \textcolor{blue}{7.39}	&	47.38	&	18.417	&	\textcolor{cyan}{2.275}&	\textcolor{cyan}{0.641}&	\textcolor{cyan}{6.968}\\
         FusionMamba & \textcolor{blue}{7.39}	&	\textcolor{cyan}{53.89}	&	17.562	&	\textcolor{red}{2.588}	&	0.597	&	6.504\\
        EMMA & \textcolor{cyan}{7.38}	&	46.02	&	18.063	&	1.877	&	0.568	&	6.944\\
        
        \bf{DIFF-MF} & \textcolor{red}{7.52}	&	\textcolor{red}{71.69}	&	\textcolor{red}{22.816}	&	\textcolor{blue}{2.402}	&	\textcolor{red}{0.643}	&	\textcolor{red}{8.195}	\\
        \bottomrule
    \end{tabular}
    }
\end{table}

\noindent \textbf{Competing methods.}
We compared the $7$ state-of-the-art methods on three publicly available datasets (M$^{3}$FD~\cite{liu2022target}, TNO~\cite{toet2012progress} and DroneVehicle~\cite{sun2020drone}).  In these comparison methods, TarDAL~\cite{liu2022target} is a GAN-based method. DIDFuse~\cite{zhao2020didfuse} and CDDFuse~\cite{zhao2023cddfuse} are deep learning-based image decomposition methods. SwinFusion~\cite{ma2022swinfusion} and EMMA~\cite{zhao2024equivariant} are the Transformer-based methods. MambaDFuse~\cite{li2024mambadfuse} and FusionMamba~\cite{xie2024fusionmamba} are the Mamba-based methods.

\subsection{Evaluation on the M$^{3}$FD dataset}
\noindent \textbf{Quantitative Comparisons.}
Table~\ref{tab:compare} presents the results of the quantitative evaluation on the M$^{3}$FD dataset, where our method achieves the best in $5$ metrics and the second  best performance in the remaining metrics, respectively. Specifically, our method demonstrates significant advantages in EN, SD, AG, and SF metrics, indicating that the fused results effectively retain abundant texture details and contrast information. Meanwhile, our approach leads in the VIF metric, suggesting our fusion results are more beneficial to the visual perception effect of human eyes. Regarding the MI metric, our fusion results achieve the second best, demonstrating satisfactory capability in preserving source image information.

\noindent \textbf{Qualitative Comparisons.}
\begin{figure*}[ht]
    \centering
    \includegraphics[width=0.9\linewidth]{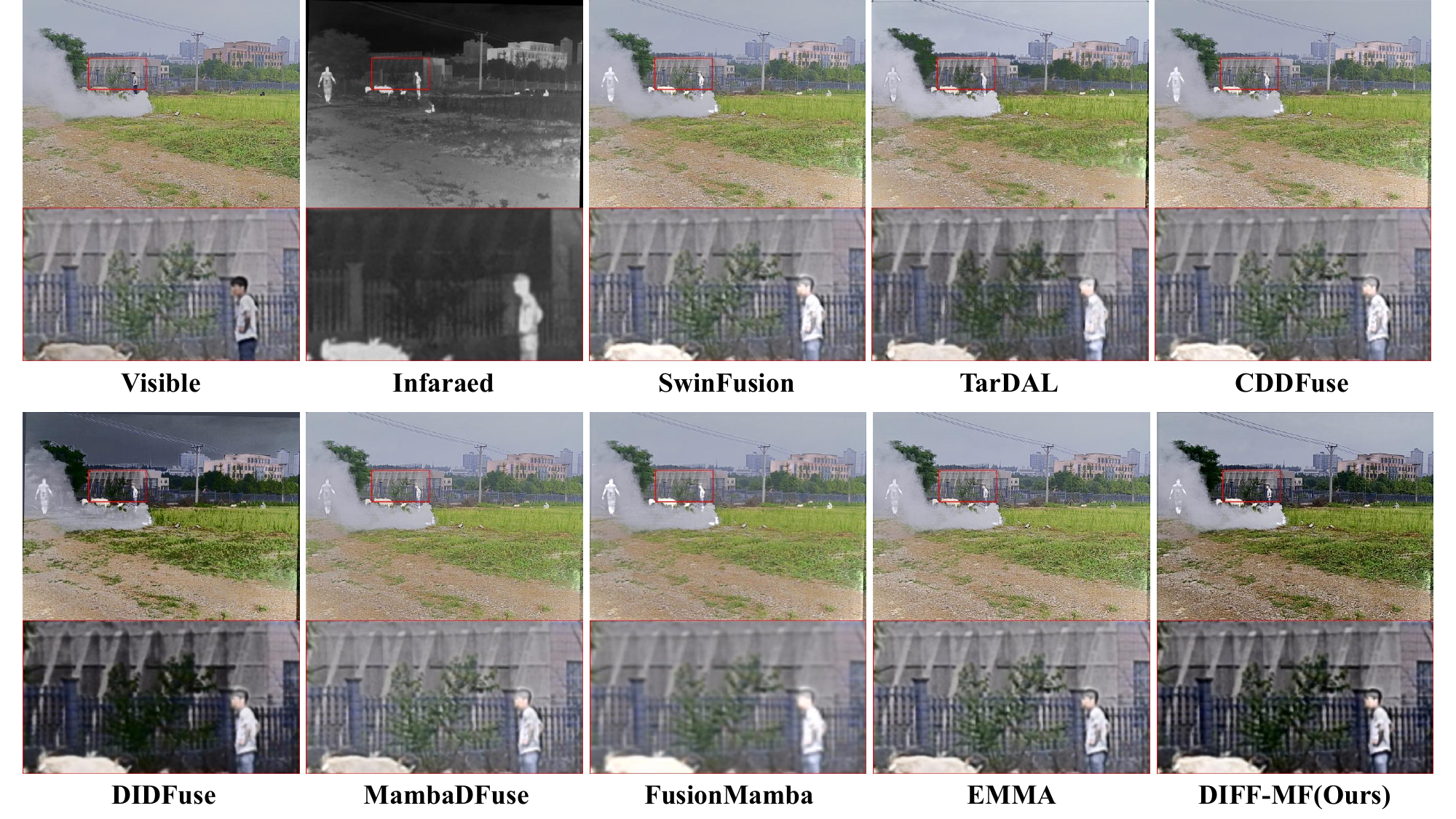}
    \caption{Qualitative comparisons of various methods on representative images selected from the M$^{3}$FD dataset.}
    \label{fig:m3fd}
\end{figure*}

To better show the superiority of our model, we assigned the color information of the 3-channel visible image to the single-channel fused image through the color space conversion between RGB and YCbCr. As shown in Fig.~\ref{fig:m3fd}, while all existing methods can reduce the interference of smoke in visible images on fused results, our method not only mitigates smoke effects effectively but also preserves a greater amount of detail and texture information in normal regions. The person in our fused images achieves a balanced enhancement in visibility while maintaining the closest resemblance to the original visible images. Furthermore, compared to other approaches, our method exhibits better performance in preserving texture details in foreground terrain and background buildings.

\subsection{Evaluation on TNO datasets}
\noindent \textbf{Quantitative Comparisons.}
Table~\ref{tab:compare} reports the performance of the different methods on the TNO dataset for $6$ metrics. Our method achieves the best results in $4$ metrics. In particular, our fused results maintain strong performance in the SD, SF, and AG metrics, demonstrating sufficient detail our fused outputs have abundant details. As our results attain the second best performance in MI and the highest score in VIF among compared methods, it means effective preservation of valuable source information and better results with human visual perception.

\begin{figure*}[ht]
  \centering
  \includegraphics[width=0.9\linewidth]{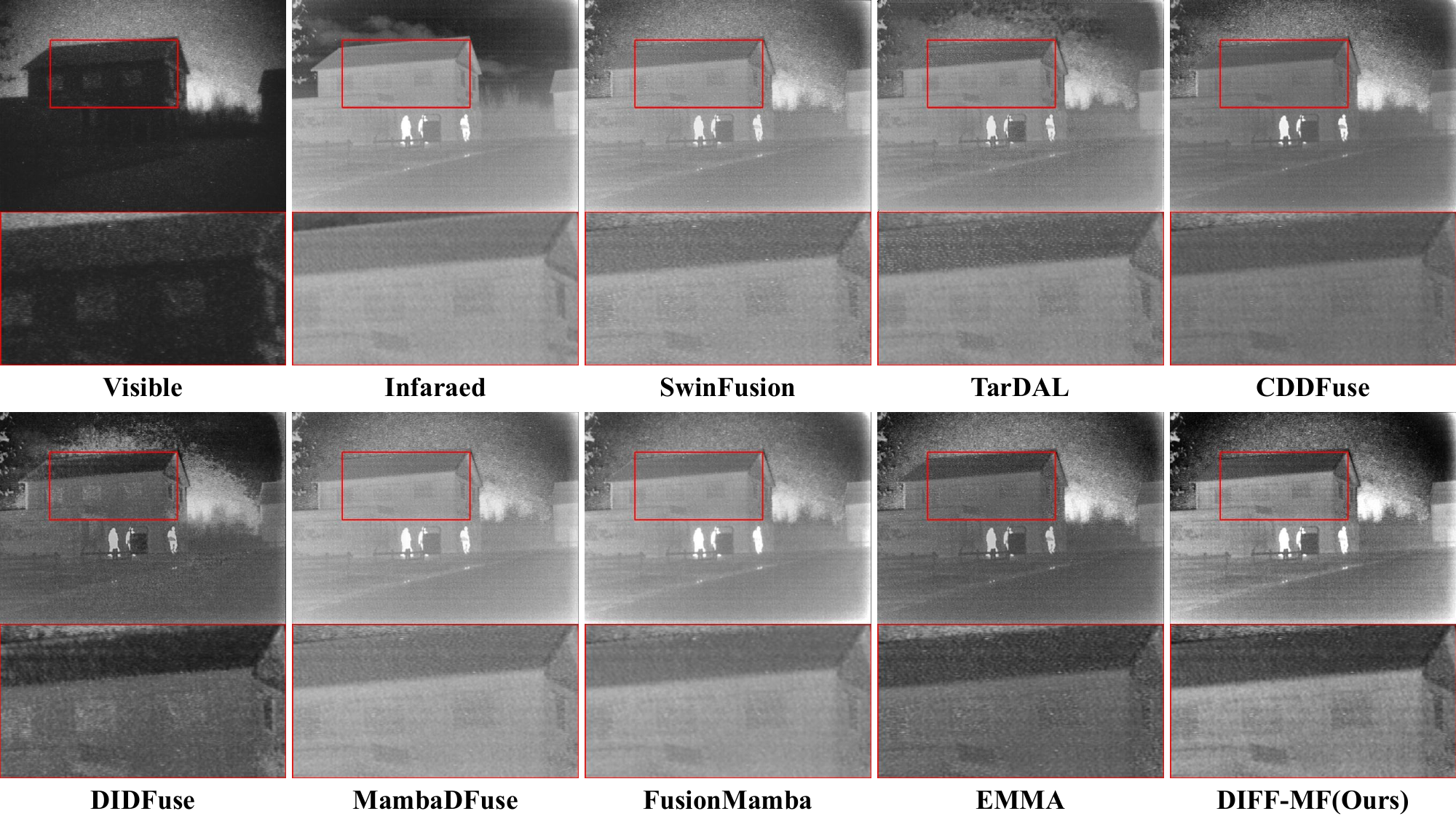}
  \caption{Qualitative comparisons of various methods on represen
tative images selected from the TNO dataset.}
  \label{fig:tno}
\end{figure*}

\noindent \textbf{Qualitative Comparisons.}
As demonstrated in Fig.~\ref{fig:tno}, our fused images maintain superior detail preservation of contours compared to alternative methods under low-light conditions. Windows of houses and eaves are more clear in our fused results, while foreground features including roads and fences have clear outlines. Furthermore, the luminance distribution in our fused images achieves better alignment with human visual perception than comparative approaches.

\subsection{Evaluation on DroneVehicle datasets}
\noindent \textbf{Quantitative Comparisons.}
\begin{figure*}
    \centering
    \includegraphics[width=1\linewidth]{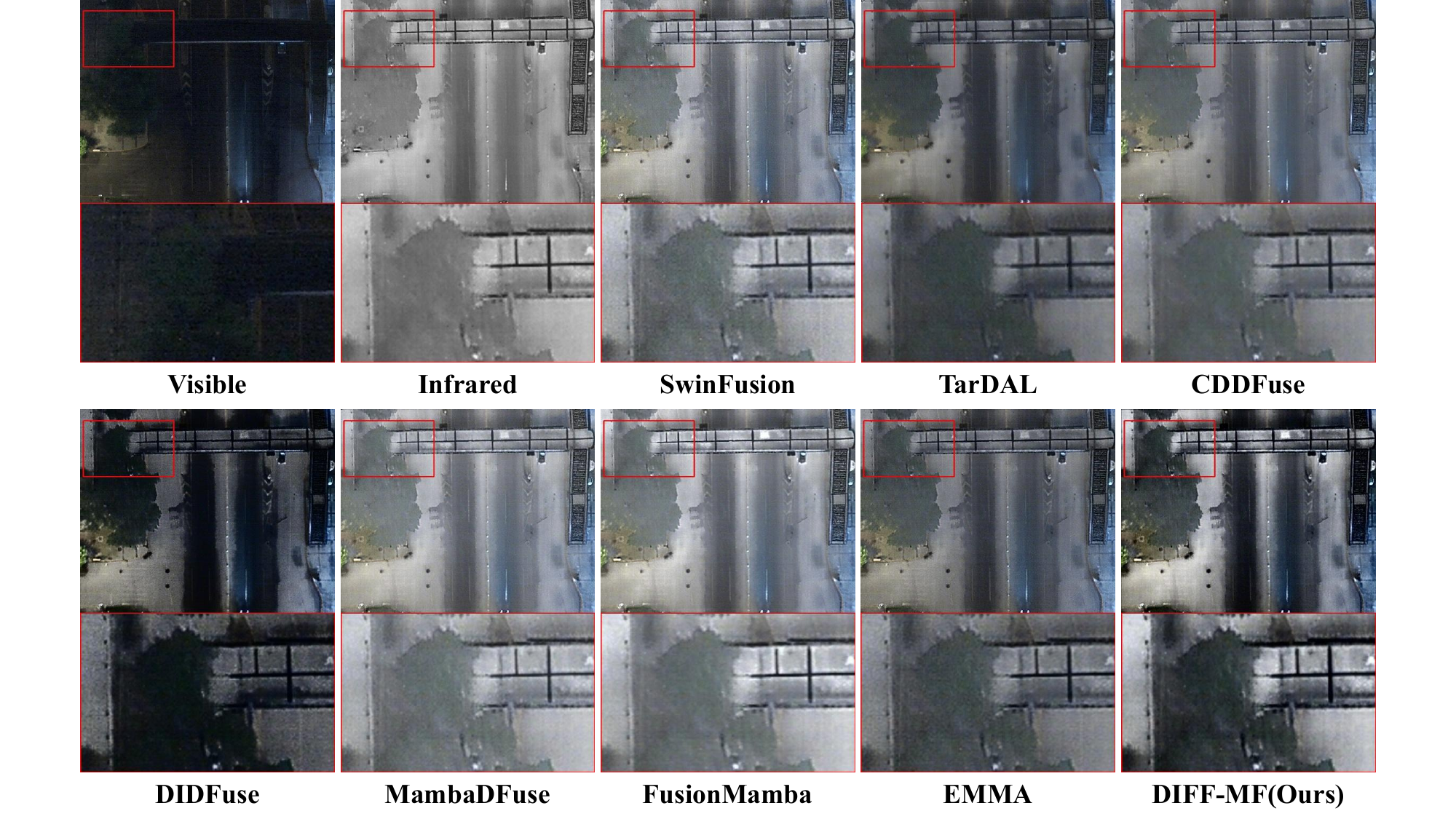}
    \caption{Qualitative comparisons of various methods on representative images selected from the DroneVehicle dataset.}
    \label{fig:drone}
\end{figure*}

The quantitative results of the different methods on the DroneVehicle dataset are reported in Table~\ref{tab:compare}. Our method outperforms all the compared methods on $5$ metrics and achieved the second  best results on the remaining metric, respectively. Our method achieves the highest score in SD, AG, and SF metrics, indicating effective preservation of substantial texture and edge information in large-scale UAV scenes. While our method is a little bit better than other methods in EN and VIF metrics, these results suggest that our methods can preserve details with human visual perception characteristics. The second-highest MI score further confirms the method's capability in retaining  source image information.

\noindent \textbf{Qualitative Comparisons.}
As shown in Fig.~\ref{fig:drone}, our fused images demonstrate enhanced clarity and superior texture detail preservation relative to existing methods. Magnified views further indicate that while comparative approaches exhibit color noise and ISO noise, our fusion results demonstrate minimal ISO noise presence.

\subsection{Ablation Study}
We conducted ablations studies on the M$^{3}$FD dataset and reported the results in Table~\ref{tab:ablation}.

\begin{table}[t]
    \centering
    \caption{Ablation Studies on M$^{3}$FD Dataset}
    \label{tab:ablation}
    \resizebox{0.5\textwidth}{!}{
    \begin{tabular}{lcccccc}
        \toprule
        \multicolumn{7}{c}{Datasets: M$^{3}$FD  Dataset}\\
        Methods & EN↑ & SD↑ & SF↑ & MI↑ & VIF↑ &  AG↑  \\
        \midrule
        w/o feature extract	& 5.3	&	12.48	&	7.487	&	2.138	&		0.484	&			2.448	 \\
        w/o channel-exchange  &  6.8	&	35.18	&	13.712	&	\textcolor{red}{3.347}	&		\textcolor{cyan}{0.806} & 4.506\\
        w/o spatial-exchange &\textcolor{cyan}{6.82}	&	\textcolor{cyan}{35.46}	&	\textcolor{blue}{14.16}	&	\textcolor{blue}{3.125} & \textcolor{red}{0.853} & \textcolor{blue}{4.736} \\
        w/o difference guidance & \textcolor{blue}{6.84}	&	\textcolor{blue}{43.49}	&	\textcolor{cyan}{13.745}	&	2.854 & 0.738 & \textcolor{cyan}{4.712} 	\\
        \bf{DIFF-MF} &  \textcolor{red}{7.19}	&	\textcolor{red}{55.09}	&	\textcolor{red}{18.787}	&	\textcolor{cyan}{3.04} & \textcolor{blue}{0.825} & \textcolor{red}{6.33}\\
        \bottomrule
    \end{tabular}
    }
\end{table}

\begin{table}[t]
    \centering
    \caption{Difference Guidance Mode Studies on M$^{3}$FD Dataset}
    \label{tab:DGM}
    \resizebox{0.49\textwidth}{!}{
    \begin{tabular}{lcccccc}
        \toprule
        \multicolumn{7}{c}{Datasets: M$^{3}$FD  Dataset}\\
        Methods & EN↑ & SD↑ & SF↑ & MI↑ & VIF↑ &  AG↑  \\
        \midrule
        diff guide v1 & 7.0	&	41.59	&	15.056	&	3.059 & 0.77 & 5.06 \\
        diff guide v2 & {7.12}	&	{53.12}	&	{17.567}	&	2.434	&	0.646&	{5.936}\\
        \bf{DIFF-MF} &  \textcolor{red}{7.19}	&	\textcolor{red}{55.09}	&	\textcolor{red}{18.787}	&	 \textcolor{red}{3.04} & \textcolor{red}{0.825} & \textcolor{red}{6.33}\\
        \bottomrule
    \end{tabular}}
\end{table}

\noindent\textbf{Difference Driven Feature Extraction.} To verify the effectiveness of difference driven feature extraction module, we first remove the overall feature extraction module from DIFF-MF. As shown in Table~\ref{tab:ablation}, all the metrics show a significant decrease after removing feature extraction, indicating that difference driven feature extraction is can extract features effectively. Then we removed the difference guidance blocks by adding the output feature maps of weighted-share dual branches back to the weighted-dependent dual branches, we found that all the metrics decrease greatly.

\noindent\textbf{Channel-Exchange Module.} We removed the channel-exchange module from the DIFF-MF to validate the importance of channel exchanging process. After removal, a decline occurs in $5$ metrics. Though the MI performs even better, the models cannot achieve a good balance.

\noindent\textbf{Spatial-Exchange Module.} We replaced the spatial-exchange module with a simple averaging method to verify the necessity of the spatial-exchange operation. The results in Table~\ref{tab:ablation} indicate performance degradation across four evaluation metrics. Although MI and VIF metrics perform better to some extent, the degradation on other metrics is considerable.

\subsection{Qualitative comparisons of Ablation Study.}
\begin{figure*}[ht]
  \centering
  \includegraphics[width=\linewidth]{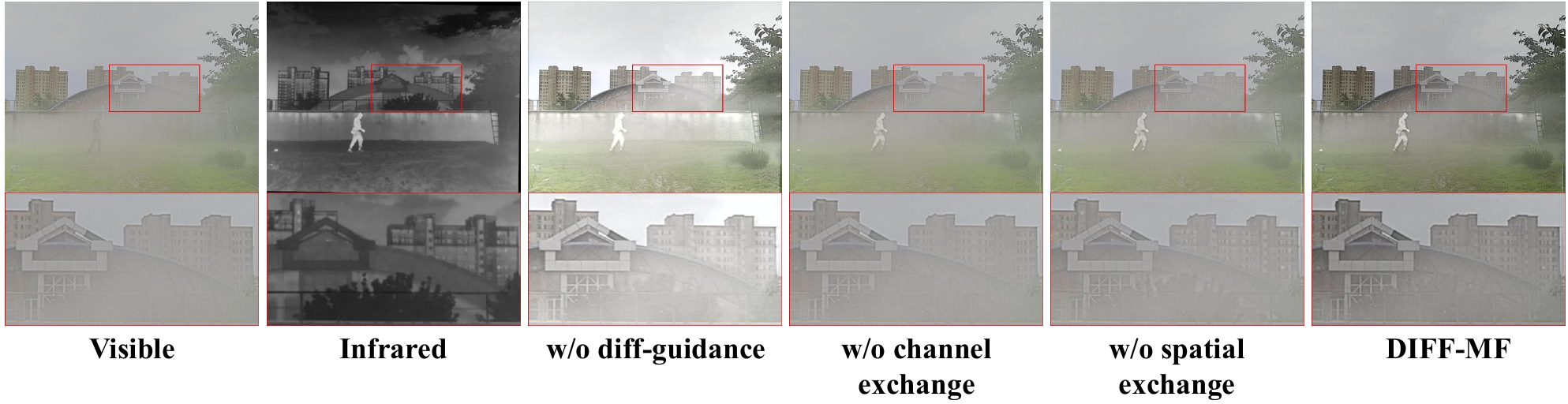}
  \caption{Qualitative comparisons of ablation study on representative images selected from the M$^{3}$FD dataset. }
  \label{fig:wo}
\end{figure*}

In this section, we present a qualitative analysis of the fusion results from our ablation study to evaluate the impact of different model components. As the results of the model without feature extraction have poor performance, we remove it form this section. As shown in Fig.~\ref{fig:wo}, only DIFF-MF have the best performance while model without difference guidance have excessive contrast losing detail texture information (person in the image). And the results form model without channel-exchange module or without spatial-exchange module is not as clear as the image from DIFF-MF.

\subsection{Analysis and Discussion}
\noindent\textbf{Difference Guidance Mode.} We changed the guidance mode for further research. Difference guidance version one (diff guide v1) will rewight the feature maps of the visible branch first and then the infrared one, while version two (diff guide v2) will reweight the feature maps in parallelly. The results are shown in Table~\ref{tab:DGM}, only the version DIFF-MF used performs best indicating that the sequence of the reweighting is related with the fusion results.

\begin{table}[t]
    \centering
    \caption{Channel-Exchange Module Ablation Studies on M$^{3}$FD Dataset}
    \label{tab:CEM}
    \resizebox{0.46\textwidth}{!}{
    \begin{tabular}{lcccccc}
        \toprule
        \multicolumn{7}{c}{Datasets: M$^{3}$FD  Dataset}\\
        Methods & EN↑ & SD↑ & SF↑ & MI↑ & VIF↑ &  AG↑  \\
        \midrule
        
        w/o SSD exchange & 6.79	&	35.03	&	13.087	&	\textcolor{red}{3.237}	&	0.78 & 4.38	 \\
        w/o channel reweighting  & 7.01	&	43.04	&	15.432	&	3.032	&	0.759 &	{5.136}\\
        \bf{DIFF-MF} &  \textcolor{red}{7.19}	&	\textcolor{red}{55.09}	&	\textcolor{red}{18.787}	&	 {3.04} & \textcolor{red}{0.825} & \textcolor{red}{6.33}\\
        \bottomrule
    \end{tabular}
}
\end{table}

\begin{table}[t]
    \centering
    \caption{Quantitative results of different SSD exchange versions}
    \label{tab:att}
    \resizebox{0.44\textwidth}{!}{
    \begin{tabular}{lcccccc}
        \toprule
        \multicolumn{7}{c}{Datasets: M$^{3}$FD  Dataset}\\
        Methods & EN↑ & SD↑ & SF↑ & MI↑ & VIF↑ &  AG↑  \\
        \midrule
        version 1 & 6.81	&	35.39	&	13.588	&	3.224	&	0.78 &	4.486\\
        version 2 & 6.81	&	35.38	&	13.604	&	\textcolor{red}{3.283}	&	0.797 &	4.523\\
        \bf{DIFF-MF} &  \textcolor{red}{7.19}	&	\textcolor{red}{55.09}	&	\textcolor{red}{18.787}	&	 {3.04} & \textcolor{red}{0.825} & \textcolor{red}{6.33}\\
        \bottomrule
    \end{tabular}}
\end{table}

\noindent\textbf{SSD Exchange and Channel Reweighting.}We also made a further study on the cross modal SSD exchange and channel reweighting, the results shown in Table~\ref{tab:CEM} demonstrate that cross modal SSD exchange have a significant impact on the fusion results and channel reweighting is also important. 

\noindent\textbf{1) SSD Exchange Mode.} As SSD Exchange can be seen as a dual form of the cross attention of the Transformer block, we change the tokens exchanged between two modals. SSD Exchange version $1$ is formulated as follows:
\begin{equation}
\begin{gathered}
{{F}_{vi}^{e'}}={{CrossMerge}}({ SSM}({ Projection}[x_{{vi}},{{B_{vi}}},{{C_{vi}}}]),\\
{{F}_{ir}^{e'}}={{CrossMerge}}({ SSM}({ Projection}[x_{{ir}},{{B_{vi}}},{{C_{vi}}}]).
\end{gathered}
\end{equation}
In this process, we only integrate the tokens from visible modal with tokens form infrared modal. In contrast, version $2$ only integrate the tokens from infrared modal with tokens form visible modal:
\begin{equation}
\begin{gathered}
{{F}_{vi}^{e'}}={{CrossMerge}}({ SSM}({ Projection}[x_{{vi}},{{B_{ir}}},{{C_{ir}}}]),\\
{{F}_{ir}^{e'}}={{CrossMerge}}({ SSM}({ Projection}[x_{{ir}},{{B_{ir}}},{{C_{ir}}}]).
\end{gathered}
\end{equation}
Table ~\ref{tab:att} shows the quantitative results of the different cross modal SSD Exchange versions.
Only the version of DIFF-MF exchanging both visible modal and infrared modal have the best performance. This may suggest that mutual exchanging can make single modal learn more about the information from another one.

\noindent\textbf{2) Residual Connection of Channel Reweighting Block.} We make a further study on the channel reweighting block. We have removed the corresponding residual connection reweighting procedure from Eq.~\ref{eq:res}. 
The Table ~\ref{tab:chmres} shows the results of the study, all metrics except the MI have a significant decrease which means our residual connection have a considerable impact on the image fusion results.

\begin{figure*}[t]
  \centering
  \includegraphics[width=0.8\textwidth]{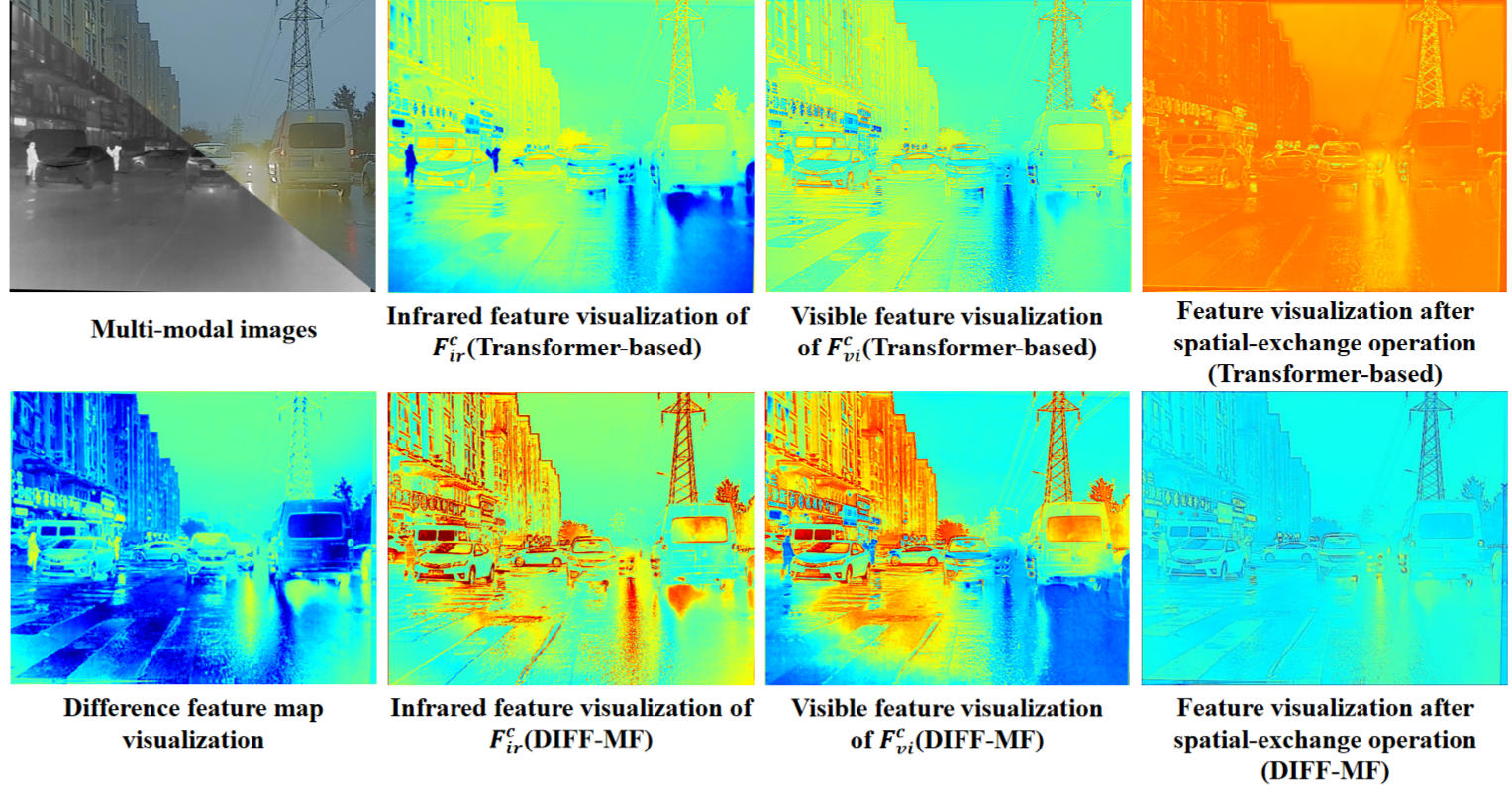}
  \caption{Feature visualization of selected images from the M$^{3}$FD dataset. }
  \label{fig:all_visual}
\end{figure*}

\begin{figure*}[t]
  \centering
  \includegraphics[width=0.85\textwidth]{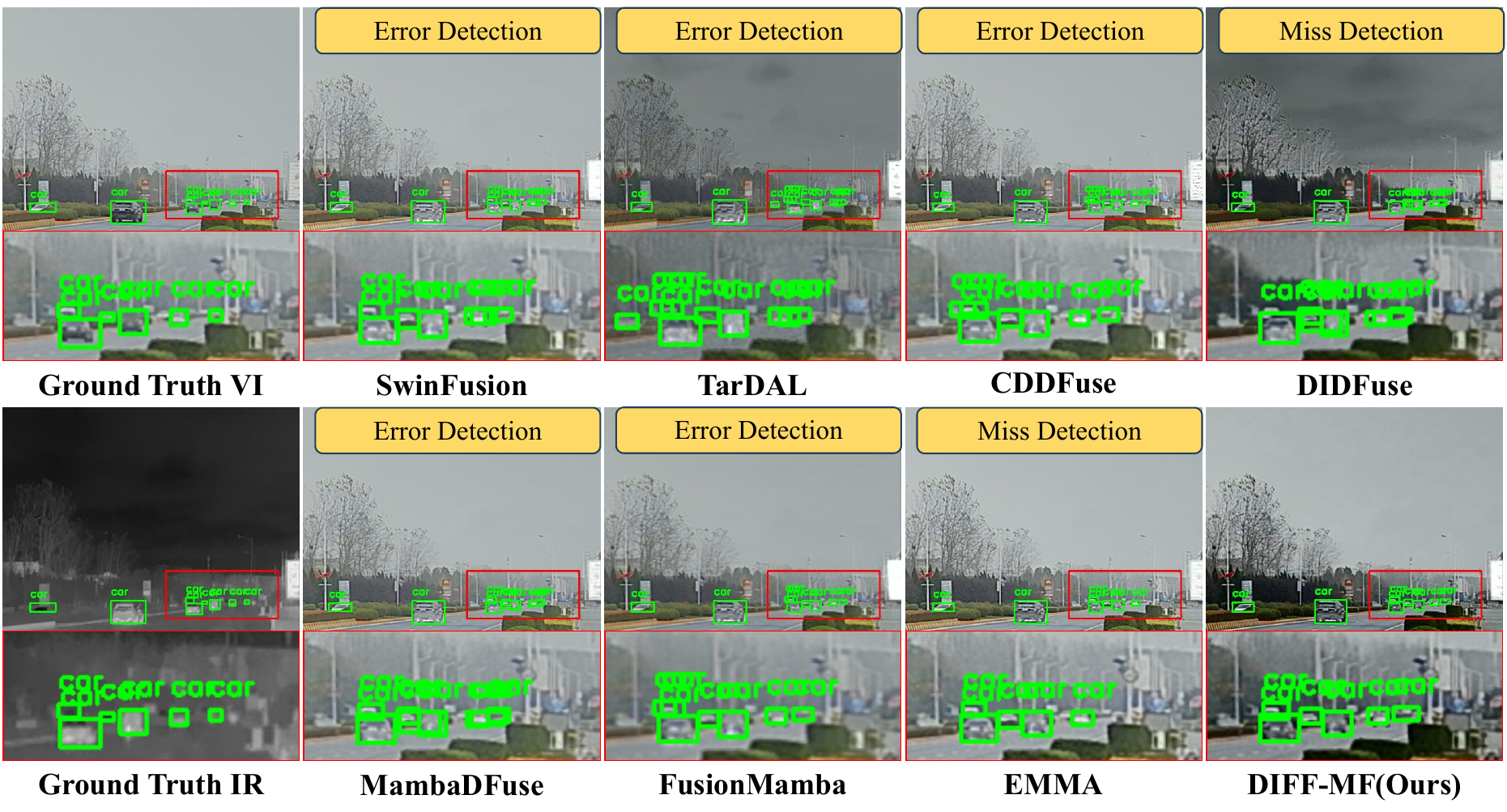}
  \caption{Qualitative comparisons of various methods on representative images selected from the M$^{3}$FD dataset. }
  \label{fig:det}
\end{figure*}

\begin{table}[t]
    \centering
    \caption{Study on Residual connection of Channel Reweighting Block}
    \label{tab:chmres}
    \resizebox{0.9\linewidth}{!}{
    \begin{tabular}{lcccccc}
        \toprule
        \multicolumn{7}{c}{Datasets: M$^{3}$FD  Dataset}\\
        Methods & EN↑ & SD↑ & SF↑ & MI↑ & VIF↑ &  AG↑  \\
        \midrule
        w/o residual connection & 6.78	&	34.49	&	13.189	&	\textcolor{red}{3.249}	&	0.785 &	4.38\\
        \bf{DIFF-MF} &  \textcolor{red}{7.19}	&	\textcolor{red}{55.09}	&	\textcolor{red}{18.787}	&	 {3.04} & \textcolor{red}{0.825} & \textcolor{red}{6.33}\\
        \bottomrule
    \end{tabular}}
\end{table}

\begin{table}[t]
    \centering
    \caption{Study on FLOPs of DIFF-MF}
    \resizebox{\linewidth}{!}{
    \begin{tabular}{lccc}
    \toprule
    Methods & input size & channel size & FLOPs \\
    \midrule
    DIFF-MF using pure transformer  & (1,1,512,512) & 32 & 336852.764G\\
    DIFF-MF using SSM  & (1,1,512,512) & 32 & 393.871G\\
    \bottomrule
    \end{tabular}}
    \label{tab:FLOPs}
\end{table}

\begin{table}[!t]

    \centering
    \caption{Downstream Task Object Detection on M$^{3}$FD Dataset}
    \label{tab:det}
    \resizebox{0.5\textwidth}{!}{
    \begin{tabular}{lccccccc}
        \toprule
        \multicolumn{8}{c}{Datasets: M$^{3}$FD  Dataset}\\
        Methods &  people   &    car     &     bus   &   lamp  &   motocycle    &   truck   &    mAP@0.5 \\
        \midrule
        SwinFusion & 0.751	&	0.8767	&	0.8682	&	0.838	&	0.7469	&	0.837	&	0.8196\\
        TarDAL & \textcolor{blue}{0.7588}	&	0.8768	&	0.9004	&	\textcolor{blue}{0.8605}	&	0.7422	&	0.839	&	0.8296\\
        DIDFuse & 0.7505	&	0.8773	&	0.8899	&	0.8414	&	0.7083	&	0.8577	&	0.8209\\
        CDDFuse & 0.7451	&	\textcolor{blue}{0.8853}	&	0.8954	&	0.8469	&	0.7319	&	\textcolor{red}{0.8681}	&	0.8288\\
        MambaDFuse & \textcolor{red}{0.7597}	&	0.8767	&	\textcolor{blue}{0.9008}	&	0.8529	&	0.7067	&	0.8228	&	0.8199\\
        FusionMamba & 0.7575	&	0.8759	&	0.8827	&	\textcolor{red}{0.8644}	&	\textcolor{red}{0.7733}	&	0.8358	&	\textcolor{blue}{0.8316}\\

        EMMA & 0.7372	&	\textcolor{red}{0.886}	&	0.8989	&	0.8526	&	\textcolor{blue}{0.7599}	&	0.8507	&	0.8309\\

        \bf{DIFF-MF} & 0.7489	&	\textcolor{blue}{0.8853}	&	\textcolor{red}{0.9021}	&	0.8389	&	0.7533	&	\textcolor{blue}{0.8652}	&	\textcolor{red}{0.8323}\\
        \bottomrule
    \end{tabular}
}
\vspace{-10pt}
\end{table}

\noindent\textbf{Analysis on FLOPs.} We make an analysis on the FLOPs of our models. Considering the size of input features is (1,1,512,512) and the embedded dimensions is set as 32, we replace all the VSS blocks in our model with the Restormer~\cite{zamir2022restormer} blocks which is a pure transformer-based model in the computer vision area and its FLOPs is less than the traditional Transformer~\cite{vaswani2017attention} blocks. The FLOPs of the attention mechanism can be formulated as $4B{(HW)^2}D$, where $B$ represents the batch size and $D$ represents the embedded dimensions. The $H,W$ is the height and width of the input features. The FLOPs of the VSS blocks can be computed as $4\times BHWDN$ where $4$ means the 4 scan routes of the VSS blocks and $N$ is the hidden state of the SSM. We set $N$ as $16$ in our models and the numbers of head in Restormer blocks is also set as $16$. Take the FLOPs of other models in the network , the final comparison of the FLOPs is shown in Table~\ref{tab:FLOPs}.It is obvious that the FLOPs of the model using SSM is much less than the model using Restormer. The attention mechanism of Restormer comsume almost $99.8\%$ of the whole models.

\noindent\textbf{Features Visualization.} We conducted feature visualization of the feature difference maps. As shown in Fig~\ref{fig:all_visual}, the difference maps effectively highlight the discrepancies between visible and infrared modality features. Key elements such as people, doorplate lamps, and other light sources are distinctly emphasized in these difference maps. Such emphasis helps the models focus more on these salient regions during the difference-guided learning process. We also present visualizations of the feature maps generated by the channel-exchange module. Fig~\ref{fig:all_visual} reveals a distinct focus: the infrared branch predominantly captures local details such as edges and textures, while the visible branch emphasizes the global context and overall structure. This clear division of representation underscores the complementary of the two modalities. Compared to the Transformer-based method in which we replaced all the VSS blocks with pure transformer blocks, DIFF-MF can capture more details both locally and globally.

In addition, we visualized the feature maps after the spatial-exchange operation. As shown in Fig~\ref{fig:all_visual}, both models are capable of capturing global features. However, whereas the pure transformer-based model retains some local information, our DIFF-MF model preserves richer fine-grained details, such as the textures of background buildings and vehicles on the road, thereby showcasing the superiority of its difference-aware paradigm.

\begin{table*}[!t]
    \centering
    \caption{Downstream Task Segmentation on FMB Dataset}
    \label{tab:seg}
    \resizebox{\linewidth}{!}{
    \begin{tabular}{lcccccccccccccccc}
        \toprule
        \multicolumn{16}{c}{Datasets: FMB  Dataset}\\
        Methods &     Unl & Road     &     Sidewalk   &   Building  &   Traffic light    &   Traffic sign   &    Plant & Sky & People & Car & Truck & Bus & Motorcycle & Cycle & Pole & MIoU \\
        \midrule
        SwinFusion & 0.214	&	0.832	&	0.532	&	0.762	&	0.75	&	0.705	&	0.827	&	0.915	&	0.508	&	0.733	&	0.897	&	0.888	&	0.907	&	0.993	&	0.365	&	0.722\\

        TarDAL & 0.191	&	0.827	&	0.545	&	0.711	&	0.76	&	0.682	&	0.826	&	0.902	&	\textcolor{blue}{0.516}	&	0.725	&	0.909	&	0.878	&	0.902	&	0.993	&	0.307	&	0.712\\

        DIDFuse & 0.207	&	\textcolor{red}{0.846}	&	\textcolor{blue}{0.596}	&	0.749	&	\textcolor{red}{0.774}	&	\textcolor{red}{0.711}	&	0.827	&	0.896	&	0.5	&	0.745	&	\textcolor{red}{0.917}	&	0.894	&	0.908	&	0.993	&	0.385	&	0.73\\
        CDDFuse & 0.213	&	0.842	&	0.54	&	0.761	&	0.747	&	\textcolor{blue}{0.706}	&	\textcolor{blue}{0.831}	&	0.91	&	0.489	&	0.741	&	0.903	&	0.894	&	0.898	&	0.993	&	0.355	&	0.721\\
        MambaDFuse & 0.213	&	\textcolor{blue}{0.844}	&	0.593	&	0.764	&	\textcolor{blue}{0.771}	&	0.696	&	0.827	&	0.916	&	0.505	&	0.745	&	0.91	&	0.887	&	\textcolor{blue}{0.909}	&	0.993	&	0.385	&	0.731\\

        FusionMamba &  0.209	&	0.842	&	0.559	&	0.762	&	0.755	&	0.7	&	0.83	&	0.909	&	\textcolor{blue}{0.516}	&	\textcolor{blue}{0.746}	&	0.904	&	0.882	&	0.906	&	0.993	&	0.384	&	0.726\\

        EMMA & \textcolor{blue}{0.22}	&	0.841	&	0.579	&	\textcolor{red}{0.789}	&	0.754	&	\textcolor{red}{0.711}	&	\textcolor{red}{0.833}	&	\textcolor{blue}{0.918}	&	0.474	&	\textcolor{red}{0.75}	&	0.905	&	\textcolor{blue}{0.899}	&	\textcolor{red}{0.91}	&	0.993	&	\textcolor{blue}{0.399}	&	\textcolor{blue}{0.732}\\

        \bf{DIFF-MF} & \textcolor{red}{0.297}	&	0.842	&	\textcolor{red}{0.612}	&	\textcolor{blue}{0.78}	&	0.767	&	\textcolor{red}{0.711}	&	\textcolor{blue}{0.831}	&	\textcolor{red}{0.919}	&	\textcolor{red}{0.527}	&	\textcolor{blue}{0.746}	&	\textcolor{blue}{0.913}	&	\textcolor{red}{0.903}	&	0.905	&	0.993	&	\textcolor{red}{0.408}	&	\textcolor{red}{0.744}\\
        \bottomrule
    \end{tabular}
    }
\end{table*}

\subsection{Downstream Task Evaluation}

\begin{figure*}[!t]
  \centering
  \includegraphics[width=0.9\linewidth]{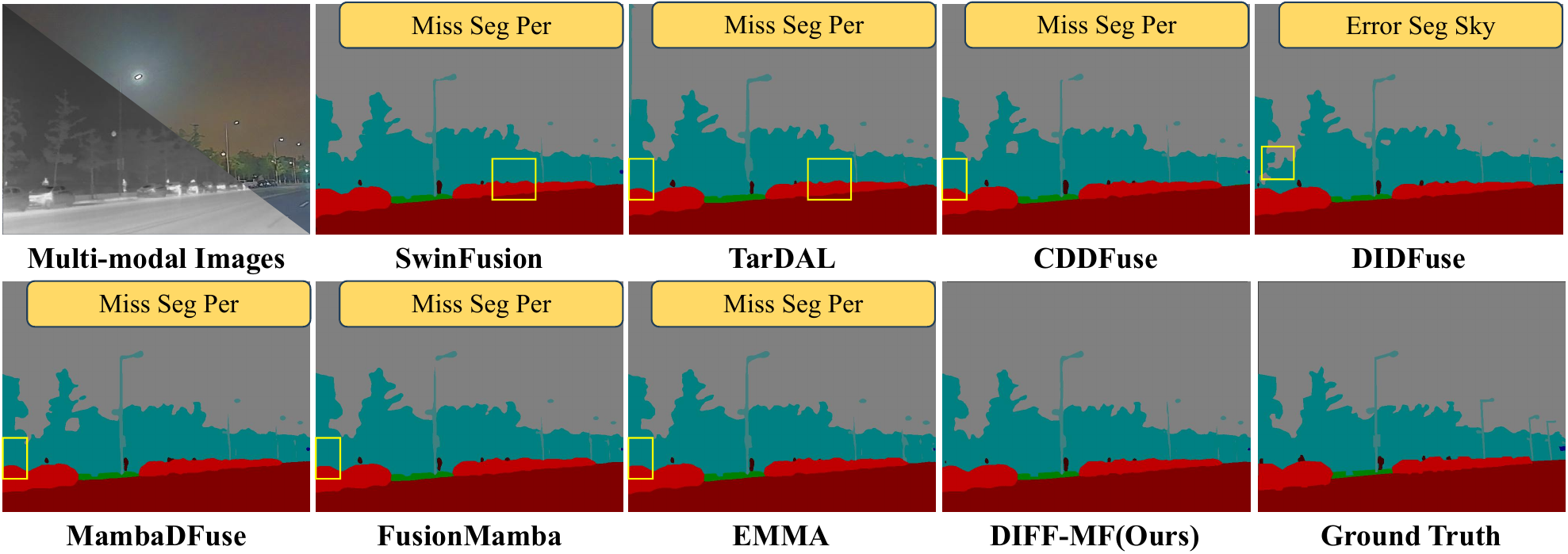}
  \caption{Qualitative comparisons of various methods on representative images selected from the FMB dataset. }
  \label{fig:seg}
\end{figure*}

To further study the performance of fusion results in downstream vision tasks, We applied the fused images to multi-modal object detection and multi-modal semantic segmentation. To ensure fairness, we individually re-train the network for each task using fusion results obtained from their own methods.

\noindent\textbf{Object Detection.}
We conducted the multi-modal object detection task on the M$^{3}$FD dataset encompassing six categories of labels: people, cars, buses, motorcycles, trucks, and lamps. We split the dataset into training/validation/test sets in an $8:1:1$ ratio. We employed YOLOv5~\cite{jocher2020ultralytics} detector to evaluate the detection performance with the metric mAP@0.5~\cite{everingham2015pascal}. The training epoch is $400$, batch size is $8$. The optimizer and initial learning rate are set as an SGD optimizer, $1.0\times 10^{-2}$, respectively. Table~\ref{tab:det} demonstrates that DIFF-MF achieves the best detection performance. The qualitative comparison results are shown in Fig. ~\ref{fig:det}. On the results of the DIFF-MF are most closely to the Ground Truth, while other results have wrong or missing detection of the car.

\noindent\textbf{Semantic Segmentation.}
We conducted the MMSS task on the FMB~\cite{liu2023multi} dataset, which consists of 1500 pairs of multi-modal images and encompasses fifteen categories of pixel-level labels: backgrounds, roads, buildings, traffic lights, traffic signs, and so on. We choose DeeplabV3+~\cite{chen2018encoder} as the segmentation network and value the performances through Intersection over Union (IoU). We separate the dataset into training/validation/test sets in an $8:1:1$ ratio. Cross-entropy loss is employed along with the SGD optimizer. The number of epochs is $300$ while the batch size and the initial learning rate are set to $4$ and $7.0\times 10^{-3}$. The segmentation results are shown in Table~\ref{tab:seg} and Fig.~\ref{fig:seg}, which demonstrate the superior performance of our method.

\section{Conclusion}
\label{sec:5}
In this paper, we present DIFF-MF, a novel difference-driven channel-spatial state space model for multimodal image fusion. By introducing a dual-branch architecture with differential guidance, our method effectively addresses the critical challenge of balancing modality-specific characteristics and cross-modal collaborative features. The proposed channel-exchange module enables adaptive reweighting of cross-modal features through state space interactions, while the multiscale spatial-scanning mechanism ensures comprehensive modeling of global contextual relationships. Unlike existing approaches that either overlook modality-specific discriminative features or inadequately model cross-scale spatial dependencies, DIFF-MF achieves synergistic integration of infrared thermal signatures and visible texture details while maintaining efficient linear computational complexity. Experimental results demonstrate that our framework successfully preserves critical information from both modalities, offering a robust solution for practical applications requiring reliable scene representation under complex imaging conditions. The proposed methodology not only advances multimodal fusion performance but also provides insights for designing efficient state-space-based architectures in vision tasks.

\bibliography{IEEEabrv}

@String(CVPR= {IEEE Conf. Comput. Vis. Pattern Recog.})

@String(ECCV= {Eur. Conf. Comput. Vis.})

@String(ICIP = {IEEE Int. Conf. Image Process.})

@String(AAAI = {AAAI})

@String(CVPR  = {CVPR})

@String(ECCV  = {ECCV})

@String(ICIP  = {ICIP})

@article{everingham2015pascal,
  title={The pascal visual object classes challenge: A retrospective},
  author={Everingham, Mark and Eslami, SM Ali and Van Gool, Luc and Williams, Christopher KI and Winn, John and Zisserman, Andrew},
  journal={International journal of computer vision},
  volume={111},
  number={1},
  pages={98--136},
  year={2015},
  publisher={Springer}
}

@inproceedings{zhao2023cddfuse,
  title={Cddfuse: Correlation-driven dual-branch feature decomposition for multi-modality image fusion},
  author={Zhao, Zixiang and Bai, Haowen and Zhang, Jiangshe and Zhang, Yulun and Xu, Shuang and Lin, Zudi and Timofte, Radu and Van Gool, Luc},
  booktitle={Proceedings of the IEEE/CVF conference on computer vision and pattern recognition},
  pages={5906--5916},
  year={2023}
}

@article{ma2022swinfusion,
  title={SwinFusion: Cross-domain long-range learning for general image fusion via swin transformer},
  author={Ma, Jiayi and Tang, Linfeng and Fan, Fan and Huang, Jun and Mei, Xiaoguang and Ma, Yong},
  journal={IEEE/CAA Journal of Automatica Sinica},
  volume={9},
  number={7},
  pages={1200--1217},
  year={2022},
  publisher={IEEE}
}

@inproceedings{zhao2024equivariant,
  title={Equivariant multi-modality image fusion},
  author={Zhao, Zixiang and Bai, Haowen and Zhang, Jiangshe and Zhang, Yulun and Zhang, Kai and Xu, Shuang and Chen, Dongdong and Timofte, Radu and Van Gool, Luc},
  booktitle={Proceedings of the IEEE/CVF conference on computer vision and pattern recognition},
  pages={25912--25921},
  year={2024}
}

@inproceedings{zamir2022restormer,
  title={Restormer: Efficient transformer for high-resolution image restoration},
  author={Zamir, Syed Waqas and Arora, Aditya and Khan, Salman and Hayat, Munawar and Khan, Fahad Shahbaz and Yang, Ming-Hsuan},
  booktitle={Proceedings of the IEEE/CVF conference on computer vision and pattern recognition},
  pages={5728--5739},
  year={2022}
}

@article{gu2023mamba,
  title={Mamba: Linear-time sequence modeling with selective state spaces},
  author={Gu, Albert and Dao, Tri},
  journal={arXiv preprint arXiv:2312.00752},
  year={2023}
}

@article{liu2024vmamba,
  title={Vmamba: Visual state space model},
  author={Liu, Yue and Tian, Yunjie and Zhao, Yuzhong and Yu, Hongtian and Xie, Lingxi and Wang, Yaowei and Ye, Qixiang and Jiao, Jianbin and Liu, Yunfan},
  journal={Advances in neural information processing systems},
  volume={37},
  pages={103031--103063},
  year={2024}
}

@article{xie2024fusionmamba,
  title={Fusionmamba: Dynamic feature enhancement for multimodal image fusion with mamba},
  author={Xie, Xinyu and Cui, Yawen and Tan, Tao and Zheng, Xubin and Yu, Zitong},
  journal={Visual Intelligence},
  volume={2},
  number={1},
  pages={37},
  year={2024},
  publisher={Springer}
}

@article{li2024mambadfuse,
  title={Mambadfuse: A mamba-based dual-phase model for multi-modality image fusion},
  author={Li, Zhe and Pan, Haiwei and Zhang, Kejia and Wang, Yuhua and Yu, Fengming},
  journal={arXiv preprint arXiv:2404.08406},
  year={2024}
}

@article{gu2021combining,
  title={Combining recurrent, convolutional, and continuous-time models with linear state space layers},
  author={Gu, Albert and Johnson, Isys and Goel, Karan and Saab, Khaled and Dao, Tri and Rudra, Atri and R{\'e}, Christopher},
  journal={Advances in neural information processing systems},
  volume={34},
  pages={572--585},
  year={2021}
}

@article{dao2024transformers,
  title={Transformers are ssms: Generalized models and efficient algorithms through structured state space duality},
  author={Dao, Tri and Gu, Albert},
  journal={arXiv preprint arXiv:2405.21060},
  year={2024}
}

@article{chen2024changemamba,
  title={Changemamba: Remote sensing change detection with spatio-temporal state space model},
  author={Chen, Hongruixuan and Song, Jian and Han, Chengxi and Xia, Junshi and Yokoya, Naoto},
  journal={IEEE Transactions on Geoscience and Remote Sensing},
  year={2024},
  publisher={IEEE}
}

@article{li2018densefuse,
  title={DenseFuse: A fusion approach to infrared and visible images},
  author={Li, Hui and Wu, Xiao-Jun},
  journal={IEEE Transactions on Image Processing},
  volume={28},
  number={5},
  pages={2614--2623},
  year={2018},
  publisher={IEEE}
}

@article{zhang2020ifcnn,
  title={IFCNN: A general image fusion framework based on convolutional neural network},
  author={Zhang, Yu and Liu, Yu and Sun, Peng and Yan, Han and Zhao, Xiaolin and Zhang, Li},
  journal={Information Fusion},
  volume={54},
  pages={99--118},
  year={2020},
  publisher={Elsevier}
}

@inproceedings{liu2022target,
  title={Target-aware Dual Adversarial Learning and a Multi-scenario Multi-Modality Benchmark to Fuse Infrared and Visible for Object Detection},
  author={Liu, Jinyuan and Fan, Xin and Huang, Zhanbo and Wu, Guanyao and Liu, Risheng and Zhong, Wei and Luo, Zhongxuan},
  booktitle={Proceedings of the IEEE/CVF Conference on Computer Vision and Pattern Recognition (CVPR)},
  pages={5802--5811},
  year={2022}
}

@article{li2021rfn,
  title={RFN-Nest: An end-to-end residual fusion network for infrared and visible images},
  author={Li, Hui and Wu, Xiao-Jun and Kittler, Josef},
  journal={Information Fusion},
  volume={73},
  pages={72--86},
  year={2021},
  publisher={Elsevier}
}

@inproceedings{zhao2020didfuse,
  author= {Zixiang Zhao and Shuang Xu and Chunxia Zhang and Junmin Liu and Jiangshe Zhang and Pengfei Li},
  title = {DIDFuse: Deep Image Decomposition for Infrared and Visible Image Fusion},
  booktitle = {Proceedings of the Twenty-Ninth International Conference on International Joint Conferences on Artificial Intelligence},
  pages = {970--976},
  publisher = {ijcai.org},
  year = {2020}
}

@article{ma2019fusiongan,
  title={FusionGAN: A generative adversarial network for infrared and visible image fusion},
  author={Ma, Jiayi and Yu, Wei and Liang, Pengwei and Li, Chang and Jiang, Junjun},
  journal={Information Fusion},
  volume={48},
  pages={11--26},
  year={2019},
  publisher={Elsevier}
}

@article{Roberts2008AssessmentOI,
  title={Assessment of image fusion procedures using entropy, image quality, and multispectral classification},
  author={John Roberts and Jan A. N. van Aardt and Fethi B. Ahmed},
  journal={Journal of Applied Remote Sensing},
  year={2008},
  volume={2}
}

@article{Eskicioglu1995ImageQM,
  title={Image quality measures and their performance},
  author={Ahmet M. Eskicioglu and Paul S. Fisher},
  journal={IEEE Trans. Commun.},
  year={1995},
  volume={43},
  pages={2959-2965}
}

@article{Han2013ANI,
  title={A new image fusion performance metric based on visual information fidelity},
  author={Yu Han and Yunze Cai and Yin Cao and Xiaoming Xu},
  journal={Information Fusion},
  year={2013},
  volume={14},
  pages={127-135}
}

@article{Qu2002InformationMF,
  title={Information measure for performance of image fusion},
  author={Guihong Qu and Dali Zhang and P. Yan},
  journal={Electronics Letters},
  year={2002},
  volume={38},
  pages={313-315}
}

@article{Cui2015DetailPF,
  title={Detail preserved fusion of visible and infrared images using regional saliency extraction and multi-scale image decomposition},
  author={Guangmang Cui and Huajun Feng and Zhi-hai Xu and Qi Li and Yue-ting Chen},
  journal={Optics Communications},
  year={2015},
  volume={341},
  pages={199-209}
}

@article{Xu2022U2FusionAU,
  title={U2Fusion: A Unified Unsupervised Image Fusion Network},
  author={Han Xu and Jiayi Ma and Junjun Jiang and Xiaojie Guo and Haibin Ling},
  journal={IEEE Transactions on Pattern Analysis and Machine Intelligence},
  year={2022},
  volume={44},
  pages={502-518}
}

@article{jocher2020ultralytics,
  title={ultralytics/yolov5: v3. 1-bug fixes and performance improvements},
  author={Jocher, Glenn and Stoken, Alex and Borovec, Jirka and Changyu, Liu and Hogan, Adam and Diaconu, Laurentiu and Ingham, Francisco and Poznanski, Jake and Fang, Jiacong and Yu, Lijun and others},
  journal={Zenodo},
  year={2020}
}

@inproceedings{chen2018encoder,
  title={Encoder-decoder with atrous separable convolution for semantic image segmentation},
  author={Chen, Liang-Chieh and Zhu, Yukun and Papandreou, George and Schroff, Florian and Adam, Hartwig},
  booktitle={Proceedings of the European conference on computer vision (ECCV)},
  pages={801--818},
  year={2018}
}

@inproceedings{liu2023multi,
  title={Multi-interactive feature learning and a full-time multi-modality benchmark for image fusion and segmentation},
  author={Liu, Jinyuan and Liu, Zhu and Wu, Guanyao and Ma, Long and Liu, Risheng and Zhong, Wei and Luo, Zhongxuan and Fan, Xin},
  booktitle={Proceedings of the IEEE/CVF international conference on computer vision},
  pages={8115--8124},
  year={2023}
}

@ARTICLE{sun2020drone,
  title={Drone-based RGB-Infrared Cross-Modality Vehicle Detection via Uncertainty-Aware Learning}, 
  author={Sun, Yiming and Cao, Bing and Zhu, Pengfei and Hu, Qinghua},
  journal={IEEE Transactions on Circuits and Systems for Video Technology}, 
  year={2022},
  volume={},
  number={},
  pages={1-1},
  doi={10.1109/TCSVT.2022.3168279}
}

@article{toet2012progress,
  title={Progress in color night vision},
  author={Toet, Alexander and Hogervorst, Maarten A},
  journal={Optical Engineering},
  volume={51},
  number={1},
  pages={010901--010901},
  year={2012},
  publisher={Society of Photo-Optical Instrumentation Engineers}
}

@article{li2020nestfuse,
  title={NestFuse: An infrared and visible image fusion architecture based on nest connection and spatial/channel attention models},
  author={Li, Hui and Wu, Xiao-Jun and Durrani, Tariq},
  journal={IEEE Transactions on Instrumentation and Measurement},
  volume={69},
  number={12},
  pages={9645--9656},
  year={2020},
  publisher={IEEE}
}

@article{wang2024mamba,
  title={Mamba YOLO: SSMs-based YOLO for object detection},
  author={Wang, Zeyu and Li, Chen and Xu, Huiying and Zhu, Xinzhong},
  journal={arXiv preprint arXiv:2406.05835},
  year={2024}
}

@inproceedings{xing2024segmamba,
  title={Segmamba: Long-range sequential modeling mamba for 3d medical image segmentation},
  author={Xing, Zhaohu and Ye, Tian and Yang, Yijun and Liu, Guang and Zhu, Lei},
  booktitle={International Conference on Medical Image Computing and Computer-Assisted Intervention},
  pages={578--588},
  year={2024},
  organization={Springer}
}

@inproceedings{lin2024mtmamba,
  title={MTMamba: Enhancing multi-task dense scene understanding by mamba-based decoders},
  author={Lin, Baijiong and Jiang, Weisen and Chen, Pengguang and Zhang, Yu and Liu, Shu and Chen, Ying-Cong},
  booktitle={European Conference on Computer Vision},
  pages={314--330},
  year={2024},
  organization={Springer}
}

@inproceedings{zhao2023ddfm,
  title={DDFM: denoising diffusion model for multi-modality image fusion},
  author={Zhao, Zixiang and Bai, Haowen and Zhu, Yuanzhi and Zhang, Jiangshe and Xu, Shuang and Zhang, Yulun and Zhang, Kai and Meng, Deyu and Timofte, Radu and Van Gool, Luc},
  booktitle={Proceedings of the IEEE/CVF International Conference on Computer Vision},
  pages={8082--8093},
  year={2023}
}

@inproceedings{zhao2021dndt,
  title={Dndt: Infrared and visible image fusion via densenet and dual-transformer},
  author={Zhao, Haibo and Nie, Rencan},
  booktitle={2021 International Conference on Information Technology and Biomedical Engineering (ICITBE)},
  pages={71--75},
  year={2021},
  organization={IEEE}
}

@article{wang2004image,
  title={Image quality assessment: from error visibility to structural similarity},
  author={Wang, Zhou and Bovik, Alan C and Sheikh, Hamid R and Simoncelli, Eero P},
  journal={IEEE transactions on image processing},
  volume={13},
  number={4},
  pages={600--612},
  year={2004},
  publisher={IEEE}
}

@article{vaswani2017attention,
  title={Attention is all you need},
  author={Vaswani, Ashish and Shazeer, Noam and Parmar, Niki and Uszkoreit, Jakob and Jones, Llion and Gomez, Aidan N and Kaiser, {\L}ukasz and Polosukhin, Illia},
  journal={Advances in neural information processing systems},
  volume={30},
  year={2017}
}

@inproceedings{ram2017deepfuse,
  title={Deepfuse: A deep unsupervised approach for exposure fusion with extreme exposure image pairs},
  author={Ram Prabhakar, K and Sai Srikar, V and Venkatesh Babu, R},
  booktitle={Proceedings of the IEEE international conference on computer vision},
  pages={4714--4722},
  year={2017}
}

@article{xu2020mef,
  title={MEF-GAN: Multi-exposure image fusion via generative adversarial networks},
  author={Xu, Han and Ma, Jiayi and Zhang, Xiao-Ping},
  journal={IEEE Transactions on Image Processing},
  volume={29},
  pages={7203--7216},
  year={2020},
  publisher={IEEE}
}

@article{zhang2021mff,
  title={MFF-GAN: An unsupervised generative adversarial network with adaptive and gradient joint constraints for multi-focus image fusion},
  author={Zhang, Hao and Le, Zhuliang and Shao, Zhenfeng and Xu, Han and Ma, Jiayi},
  journal={Information Fusion},
  volume={66},
  pages={40--53},
  year={2021},
  publisher={Elsevier}
}

@inproceedings{qu2022transmef,
  title={Transmef: A transformer-based multi-exposure image fusion framework using self-supervised multi-task learning},
  author={Qu, Linhao and Liu, Shaolei and Wang, Manning and Song, Zhijian},
  booktitle={Proceedings of the AAAI conference on artificial intelligence},
  volume={36},
  number={2},
  pages={2126--2134},
  year={2022}
}

@inproceedings{vs2022image,
  title={Image fusion transformer},
  author={Vs, Vibashan and Valanarasu, Jeya Maria Jose and Oza, Poojan and Patel, Vishal M},
  booktitle={2022 IEEE International conference on image processing (ICIP)},
  pages={3566--3570},
  year={2022},
  organization={IEEE}
}

@article{fu2021ppt,
  title={PPT fusion: Pyramid patch transformerfor a case study in image fusion},
  author={Fu, Yu and Xu, TianYang and Wu, XiaoJun and Kittler, Josef},
  journal={arXiv preprint arXiv:2107.13967},
  year={2021}
}

@article{yi2024diff,
  title={Diff-IF: Multi-modality image fusion via diffusion model with fusion knowledge prior},
  author={Yi, Xunpeng and Tang, Linfeng and Zhang, Hao and Xu, Han and Ma, Jiayi},
  journal={Information Fusion},
  volume={110},
  pages={102450},
  year={2024},
  publisher={Elsevier}
}

@article{zhu2024vision,
  title={Vision mamba: Efficient visual representation learning with bidirectional state space model},
  author={Zhu, Lianghui and Liao, Bencheng and Zhang, Qian and Wang, Xinlong and Liu, Wenyu and Wang, Xinggang},
  journal={arXiv preprint arXiv:2401.09417},
  year={2024}
}

@article{han2022survey,
  title={A survey on vision transformer},
  author={Han, Kai and Wang, Yunhe and Chen, Hanting and Chen, Xinghao and Guo, Jianyuan and Liu, Zhenhua and Tang, Yehui and Xiao, An and Xu, Chunjing and Xu, Yixing and others},
  journal={IEEE transactions on pattern analysis and machine intelligence},
  volume={45},
  number={1},
  pages={87--110},
  year={2022},
  publisher={IEEE}
}

@article{huang2025t,
  title={T 2 EA: Target-aware Taylor expansion approximation network for infrared and visible image fusion},
  author={Huang, Zhenghua and Lin, Cheng and Xu, Biyun and Xia, Menghan and Li, Qian and Li, Yansheng and Sang, Nong},
  journal={IEEE Transactions on Circuits and Systems for Video Technology},
  year={2025},
  publisher={IEEE}
}

@article{li2025graph,
  title={Graph representation learning for infrared and visible image fusion},
  author={Li, Jing and Bai, Lu and Yang, Bin and Li, Chang and Ma, Lingfei},
  journal={IEEE Transactions on Automation Science and Engineering},
  year={2025},
  publisher={IEEE}
}

@article{micheli2009neural,
  title={Neural network for graphs: A contextual constructive approach},
  author={Micheli, Alessio},
  journal={IEEE Transactions on Neural Networks},
  volume={20},
  number={3},
  pages={498--511},
  year={2009},
  publisher={IEEE}
}

@article{zhao2023interactive,
  title={Interactive feature embedding for infrared and visible image fusion},
  author={Zhao, Fan and Zhao, Wenda and Lu, Huchuan},
  journal={IEEE Transactions on Neural Networks and Learning Systems},
  volume={35},
  number={9},
  pages={12810--12822},
  year={2023},
  publisher={IEEE}
}

@article{yang2025dsfuse,
  title={DSFuse: A Dual-Diffusion Structure for Feature Fidelity Infrared and Visible Image Fusion},
  author={Yang, Zhijia and Gao, Kun and Zhang, Yanzheng and Zhang, Xiaodian and Hu, Zibo and Wang, Junwei and Wang, Jingyi and Li, Wei},
  journal={IEEE Transactions on Neural Networks and Learning Systems},
  year={2025},
  publisher={IEEE}
}

@ARTICLE{10177917,
  author={Zhao, Cheng and Yang, Peng and Zhou, Feng and Yue, Guanghui and Wang, Shuigen and Wu, Huisi and Chen, Guoliang and Wang, Tianfu and Lei, Baiying},
  journal={IEEE Transactions on Neural Networks and Learning Systems}, 
  title={MHW-GAN: Multidiscriminator Hierarchical Wavelet Generative Adversarial Network for Multimodal Image Fusion}, 
  year={2024},
  volume={35},
  number={10},
  pages={13713-13727},
  doi={10.1109/TNNLS.2023.3271059}}

@ARTICLE{10278227,
  author={Liu, Jinyang and Li, Shutao and Liu, Haibo and Dian, Renwei and Wei, Xiaohui},
  journal={IEEE Transactions on Neural Networks and Learning Systems}, 
  title={A Lightweight Pixel-Level Unified Image Fusion Network}, 
  year={2024},
  volume={35},
  number={12},
  pages={18120-18132},
  doi={10.1109/TNNLS.2023.3311820}}

@ARTICLE{9697423,
  author={Han, Qihui and Jung, Cheolkon},
  journal={IEEE Transactions on Neural Networks and Learning Systems}, 
  title={Deep Selective Fusion of Visible and Near-Infrared Images Using Unsupervised U-Net}, 
  year={2025},
  volume={36},
  number={3},
  pages={4172-4183},
  doi={10.1109/TNNLS.2022.3142780}}

@ARTICLE{10190200,
  author={Li, Jiawei and Liu, Jinyuan and Zhou, Shihua and Zhang, Qiang and Kasabov, Nikola K.},
  journal={IEEE Transactions on Neural Networks and Learning Systems}, 
  title={GeSeNet: A General Semantic-Guided Network With Couple Mask Ensemble for Medical Image Fusion}, 
  year={2024},
  volume={35},
  number={11},
  pages={16248-16261},
  doi={10.1109/TNNLS.2023.3293274}}

@ARTICLE{10026659,
  author={Xu, Guoxia and He, Chunming and Wang, Hao and Zhu, Hu and Ding, Weiping},
  journal={IEEE Transactions on Neural Networks and Learning Systems}, 
  title={DM-Fusion: Deep Model-Driven Network for Heterogeneous Image Fusion}, 
  year={2024},
  volume={35},
  number={7},
  pages={10071-10085},
  doi={10.1109/TNNLS.2023.3238511}}

@ARTICLE{10713288,
  author={Liu, Jinyang and Li, Shutao and Tan, Lishan and Dian, Renwei},
  journal={IEEE Transactions on Neural Networks and Learning Systems}, 
  title={Denoiser Learning for Infrared and Visible Image Fusion}, 
  year={2025},
  volume={36},
  number={7},
  pages={13470-13482},
  doi={10.1109/TNNLS.2024.3454811}}
\bibliographystyle{IEEEtran}

\vfill

\end{document}